\documentclass[10pt]{article} 

\usepackage[preprint]{rlj} 

\usepackage{amssymb}            
\usepackage{mathtools}          
\usepackage{mathrsfs}           
\usepackage{graphicx}           
\usepackage{subcaption}         
\usepackage[space]{grffile}     
\usepackage{url}                
\usepackage{lipsum}             

\usepackage{multicol}
\usepackage{wrapfig}

\usepackage{booktabs} 
\usepackage{verbatim}
\usepackage{amsfonts}
\usepackage{multirow}
\usepackage{algorithm}
\usepackage{algpseudocode}
\usepackage{tabularx}
\usepackage{amsthm}

\newcommand{\method}{Multitask discriminator Proximity-Guided IRL}
\newcommand{\M}{MPG} 
\newcommand{\mysetting}{FM-IRL}

\definecolor{mpirlblue}{RGB}{0, 51, 153}
\title{Generalize and Guide: Decomposing Rewards for Few-Shot Inverse Reinforcement Learning}

\setrunningtitle{Few-Shot IRL via Multi-Task Discrimination and Proximity Rewards}

\author{Ziyi Liu\textsuperscript{1,$\dagger$}, Grace Zhang\textsuperscript{1,$\dagger$}}

\emails{\ \{ziyiliu29,graciez168\}@gmail.com}

\affiliations{
$^{1}$\textbf{University of Southern California}
\par 
$^\dagger$ indicating equal contribution
}

\contribution{
      We introduce a novel problem setting: \emph{few-shot IRL with multi-task demonstrations (\mysetting)}, where an agent must learn a new task from limited demonstrations, while mainly leveraging information from demonstrations of related tasks and online agent experience.
    }
    {
Prior work on learning from limited demonstrations has primarily focused on using imitation learning or meta-learning, which either cannot leverage online policy optimization or require access to multi-task environments. In practice, although interaction with multi-task environments is often difficult or impractical in real-world settings, the agent typically  has access to two readily available sources of information: (1) expert demonstrations from related tasks, and (2) online interaction with the target task environment.
    }

\contribution{
     We propose \method~(\M), a reward decomposition framework that generates robust reward for expert distribution and provides guidance
for recovery from unseen states.
    }
    {To generalize beyond few-shot demonstrations, our key insight is twofold: demonstrations from other tasks can be leveraged to infer the expert distribution of a new task, and online interaction can be used to estimate temporal distance to this distribution, providing corrective guidance when the agent deviates.
    }

\contribution{
    We establish strong performance in the \mysetting~setting across diverse navigation and manipulation domains.
    }
    {
   We evaluate on maze navigation, block stacking, and robot manipulation tasks in FactorWorld. Our method achieves an average success rate of 81.2\%, outperforming the strongest baseline selected separately for each task by 24.7 percentage points on average. Our results demonstrate robustness under substantial variations and limited expert supervision.
    }


\summary{Inverse reinforcement learning (IRL) provides a powerful framework for learning from demonstrations. However, real-world tasks often exhibit substantial natural variations (e.g., picking up mugs with varying shapes), making it impractical to collect demonstrations that fully specify a new task under every possible scenario. In practice, while demonstrations for the target task are limited, it is often easier to obtain datasets of heterogeneous but related behaviors. 
This motivates the problem of \emph{few-shot IRL with multi-task demonstrations (\mysetting)}, where an agent must learn a new task with substantial variations from only a limited number of target-task demonstrations,  together with sufficient demonstrations of related tasks and online agent experience. 
To do so, we must both recover the expert distribution of the new task and provide guidance when the agent deviates from it.
We introduce \method~(\M), which learns two complementary reward components: (1) a \emph{generalizable} discriminator that transfers shared structure across related tasks to identify expert behavior in a new task, and (2) a proximity function that measures how far a state deviates from expert behavior and provides corrective \emph{guidance} during exploration.  
We demonstrate the effectiveness of our method on multiple challenging navigation and manipulation tasks under significant variations (e.g., object configurations, table layouts, and initial robot poses), achieving an average success rate of 81.2\%, outperforming the strongest baseline selected separately for each task by 24.7 percentage points on average. Project page and code are available at
\href{https://taichi-pink.github.io/Ziyi-Liu/mpg/index.html}{https://taichi-pink.github.io/Ziyi-Liu/mpg/index.html}.
}

\begin{document}

\makeCover  
\maketitle  

\begin{abstract}
Inverse reinforcement learning (IRL) provides a powerful framework for learning from demonstrations. However, real-world tasks often exhibit substantial natural variations (e.g., picking up mugs with varying shapes), making it impractical to collect demonstrations that fully specify a new task under every possible scenario. In practice, while demonstrations for the target task are limited, it is often easier to obtain datasets of heterogeneous but related behaviors. 
This motivates the problem of \emph{few-shot IRL with multi-task demonstrations (\mysetting)}, where an agent must learn a new task with substantial variations from only a limited number of target-task demonstrations,  together with sufficient demonstrations of related tasks and online agent experience. 
To do so, we must both recover the expert distribution of the new task and provide guidance when the agent deviates from it.
We introduce \method~(\M), which learns two complementary reward components: (1) a \emph{generalizable} discriminator that transfers shared structure across related tasks to identify expert behavior in a new task, and (2) a proximity function that measures how far a state deviates from expert behavior and provides corrective \emph{guidance} during exploration.  
We demonstrate the effectiveness of our method on multiple challenging navigation and manipulation tasks under significant variations (e.g., object configurations, table layouts, and initial robot poses), achieving an average success rate of 81.2\%, outperforming the strongest per-task baseline by an average of 24.7 percentage points. 
\end{abstract}

\section{Introduction}
\label{sec:introduction}

Reinforcement Learning (RL) provides a powerful framework for sequential decision-making, but its reliance on well-shaped and often hand-engineered reward functions remains a significant bottleneck~\citep{sutton2018rl, amodei2016concreteproblemsaisafety}.
Inverse Reinforcement Learning (IRL) \citep{ng2000irl} offers an alternative by inferring a reward function directly from expert demonstrations. 
However, many realistic settings include natural \emph{intra-task variations} (e.g., picking up mugs with different shapes and placements), where collecting sufficient demonstrations for every scenario is prohibitively expensive, limiting the applicability of traditional IRL.

Prior work on learning from limited demonstrations has primarily focused on using imitation learning (IL)~\citep{finn2017one, yu2018one} or  meta-inverse reinforcement learning (meta-IRL)~\citep{xu2019learning, yu2019meta}. 
However, IL methods struggle to generalize beyond the demonstrated state distribution, while meta-IRL typically requires access to multi-task training environments—an assumption that is often impractical in real-world settings. For example, demonstrations of robotic cleaning tasks across diverse furniture configurations may be available, but recreating the corresponding environments is difficult.
Importantly, the agent often still has access to two readily available sources of information: (1) expert demonstrations from related tasks, and (2) online interaction with the target task environment.
Motivated by this, we introduce a novel setting: \emph{few-shot IRL with multi-task demonstrations (\mysetting)} (see Table~\ref{tab:problem_settings}).

\begin{table}[th]
\vspace{-6pt}
\caption{Comparison of assumptions across related settings for learning from limited demonstrations.  
In practical robotics scenarios, although interaction with \emph{Multi-task Env.} is difficult to obtain, \emph{Multi-task Demos} are often available. 
Our setting reflects this scenario by leveraging \emph{Multi-task Demos} while requiring interaction \emph{only} with the \emph{Target-task Env.}, without access to \emph{Multi-task Env.}.}
\label{tab:problem_settings}
\centering
\small
\setlength{\tabcolsep}{4pt}
\begin{tabularx}{\columnwidth}{
>{\centering\arraybackslash}X
>{\centering\arraybackslash}X
>{\centering\arraybackslash}X
>{\centering\arraybackslash}X
>{\centering\arraybackslash}X}
\toprule
Problem Setting  & Target-task Demos & Multi-task Demos & Target-task Env. & Multi-task Env. \\
\midrule
Traditional IRL & \checkmark &  & \checkmark &  \\
\midrule
Few-Shot IL & \checkmark & \checkmark &  &  \\
\midrule
Meta-IRL & \checkmark & \checkmark & \checkmark & \checkmark \\
\midrule
\textbf{\mysetting~(Ours)} & \checkmark & \checkmark & \checkmark &  \\
\bottomrule
\end{tabularx}
\vspace{-8pt}
\end{table}

\begin{wrapfigure}{r}{0.45\textwidth} 
    \centering
    \includegraphics[width=0.44\textwidth]{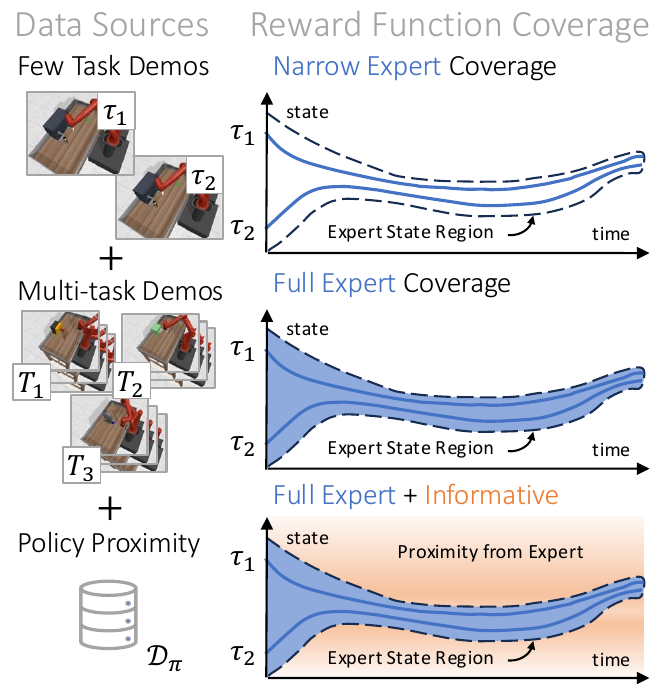}
        \caption[]{We learn a generalizeable and informative reward by making use of multi-task demonstrations and policy proximity.}

    \label{fig:teaser}
    \vspace{-15pt}
\end{wrapfigure}

This setting closely mirrors how humans learn a new task with substantial variations under limited supervision. For example, preparing a dish after being shown how just once. Beginners often (1) start by drawing on prior experience from preparing similar dishes—distilling useful behaviors and constraints that should be encouraged or avoided across related tasks. They then (2) refine their behavior through practice. When making mistakes or drifting away from behaviors consistent with the target dish (e.g., misplacing ingredients), they should return to the expert behavior. 
Through this interplay between \textbf{knowledge transfer} and \textbf{corrective feedback}, humans can master a new dish without needing demonstrations for every kitchen configuration.

Can we enable agents to similarly observe, practice, and master a new task with minimal supervision despite significant intra-task variation?
We propose \textbf{\method~(\M)}, a novel few-shot IRL method that decomposes the reward function into two synergistic components (Figure ~\ref{fig:teaser}): (1) a demonstration-conditioned multi-task discriminator that \emph{transfers knowledge} by leveraging shared structure across related tasks to generalize expert behavior to a new task under diverse variations, and (2) a proximity function that provides informative, \emph{corrective feedback} in non-expert regions, by estimating the agent’s distance to the expert state distribution. 

Overall, we identify the challenge of under-specification in realistic tasks with natural variations and propose the problem setting of \emph{\mysetting}. Towards solving this problem setting, we propose \emph{\M}, a novel method that learns a generalizable and informative reward function for effective few-shot IRL. Our experimental results on maze navigation, block stacking, and robot manipulation tasks in FactorWorld~\citep{xie2024decomposing}, demonstrate that \M~achieves state-of-the-art performance against 9 other baselines in this problem setting, outperforming the strongest per-task baseline by an average of 24.7 percentage points in success rate.

\section{Related Work}
\label{sec:related_work}






\subsection{Few-Shot Imitation Learning}
Imitation learning (IL) aims to replicate expert behavior by learning directly from expert demonstrations. 
In this paper, we distinguish IL methods as those that learn a policy directly without inferring or optimizing a reward function.  
Early approaches address few-shot IL use Behavior Cloning (BC) \citep{finn2017one, duan2017one, yu2018one}. 
\citet{hakhamaneshi2021hierarchical} extract skill models from an offline dataset to facilitate few-shot IL, while 
\citet{pmlr-v139-dance21a} learn a demonstration-conditioned policy with access to ground-truth rewards.
Other works explore offline IL~\citep{luo2023offlineoptimal, xu2022offlineimitation, chang2021offlinemitigating}, but do not explicitly address the challenge of few-shot imitation.
Overall, IL methods suffer from compounding errors  and cannot improve through online interactions without learning a reward function.
In response to this, \citet{reddy2020sqil} propose a simple, sparse reward label to allow for policy optimization through RL.   
Meanwhile, \citet{pmlr-v162-chae22a} handle environment dynamic variations by imitating multiple experts in different dynamics.

\subsection{Few-Shot Inverse Reinforcement Learning}



The most common approach to solve few-shot IRL is meta-learning, including
context-based and gradient-based methods~\citep{fu2018airl, ziebart2008maximum}. Context-based methods \citep{MHAIRL, seyed2019smile, yu2019meta} learn a latent context variable to represent tasks and meta-train a context-conditioned reward function.
Gradient-based methods ~\citep{xu2019learning} learn a good initialization for the reward function, which can be quickly adapted to a new task through a one-step gradient update. 
Meta-learning aims to enable rapid adaptation to a meta-test suite of tasks, whereas our approach focuses on sample-efficient learning for a single target task.


\citet{chen2021learning} propose DVD, a multi-task video success discriminator 
that generalize across task variations from a few robot demos but does not employ RL to learn a policy.  Similarly, ~\citet{xie2018goal} develop a success classifier for goal-conditioned tasks but do not learn a reward function.  
Our work can be viewed as an extension of these ideas. 
Other works study demonstration-efficient IRL in multi-task \citep{gleave2018multitaskmaximumentropyinverse} and multi-agent \citep{filos2021psiphi} settings.

\subsection{Proximity-based Rewards}
Popular IRL methods \citep{ho2016gail, fu2018airl} learn reward functions by discriminating between agent and expert behaviors, which may provide limited guidance in non-expert states.  
To address this, recent work introduces reward shaping mechanisms that estimate some form of proximity to the expert.  Examples includes a progress estimator for goal-conditioned tasks~\citep{lee2021generalizable}, Euclidean distance between agent's and expert's state-action pairs~\citep{hakhamaneshi2021hierarchical}, geometric distance between agent and expert distribution~\citep{dadashi2021primal,haldar2022watch}, and a transition-based reachability discriminator \citet{chiang2024expert}.
While these approaches provide useful guidance in non-expert states, they do not account for generalization across task variations with limited demonstrations.



\section{Few-Shot IRL with Multi-Task Demonstrations Problem Formulation}
\label{sec:problem}



IRL addresses the problem of learning sequential decision-making tasks from demonstrations. We consider these tasks to be Markov decision problems (MDPs) defined by the tuple $(\mathcal{S}, \mathcal{A}, \mathcal{T}, \rho, \mathcal{R})$: state space $\mathcal{S}$, action space $\mathcal{A}$, transition probabilities $\mathcal{T}$, initial state distribution $\rho$, and underlying reward function $\mathcal{R}$. We assume $\mathcal{R}$ is unkown and must instead be inferred from a set of demonstrations $\mathcal{D}$ from an expert policy $\pi^*(a|s)$. The goal is to learn a reward function $\tilde{\mathcal{R}}: \mathcal{S} \times \mathcal{A} \rightarrow \mathbb{R}$ that explains the expert behavior and can then be used to learn a policy $\pi(a|s)$.  

We focus on a \emph{few-shot} setting where the task exhibits significant \emph{intra-task variations}, arising from the initial state distribution $\rho_i$ (e.g., varying agent positions or object configurations), yet only a small set of demonstrations $\mathcal{D}_{target}$ is available. 
This limited data is insufficient for traditional IRL  to learn a reward that generalizes to unseen instances.
To overcome this challenge, we additionally provide a large multi-task demonstration dataset $\mathcal{D}_{multi} = \{\mathcal{D}_1, \mathcal{D}_2, \cdots, \mathcal{D}_T \}$. Each task $i$ has its own distinct underlying reward function $\mathcal{R}_i$, transition dynamics $\mathcal{T}$, and initial state distribution $\rho_i$ but shares the same state and action spaces $(\mathcal{S}, \mathcal{A})$.
The agent's ultimate goal is to leverage the broad knowledge in $\mathcal{D}_{multi}$, information in $\mathcal{D}_{target}$, and online samples $\mathcal{D}_{\pi}$, to learn a reward function $\tilde{\mathcal{R}}$ that enables an RL agent to learn a policy $\pi$ that successfully solves unseen instances of the target task.
\begin{figure*}[t]
   \centering
   \includegraphics[width=\textwidth]{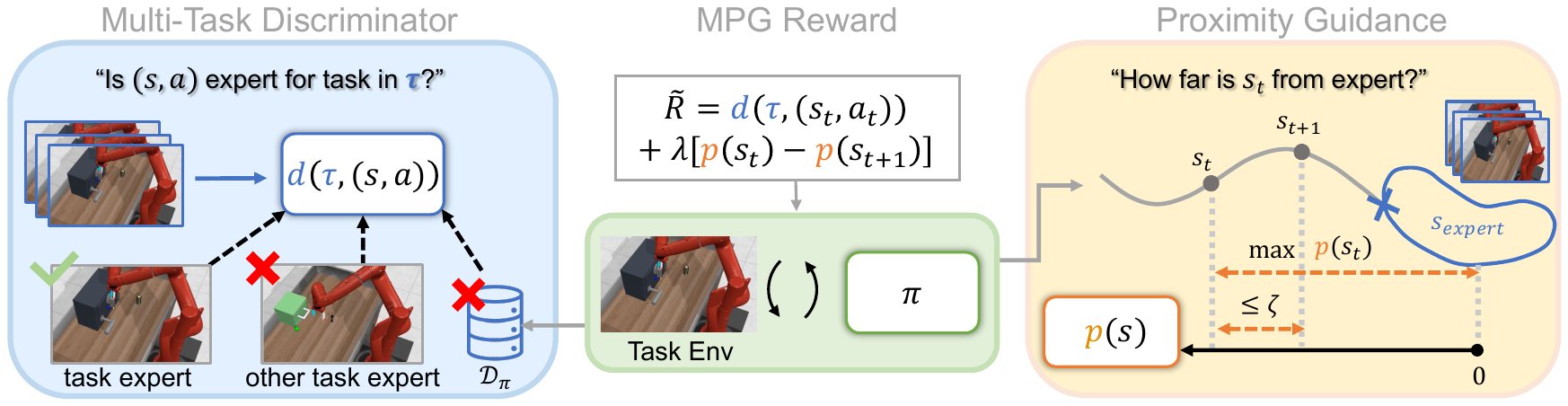}    
   \caption[]{
   Our approach learns a two-part reward function, $\tilde{R}$. The \textcolor{blue}{multi-task discriminator} extracts common structure across tasks, by predicting whether a state-action pair $(s^{i}, a^{i})$ is expert for the task in demonstration $\tau^{j}$ ($i$ and $j$ index tasks), to approximate expert distribution. 
   The \textcolor{orange}{proximity reward} estimates a state's proximity to the expert states by maximizing proximity $p_\theta(s_t)$, constrained by the triangle inequality $p_\theta(s_t) \leq p_\theta(s_{t+1}) + \zeta$.  
   Finally, the combined reward $\tilde{R}$ integrates expert recognition and proximity guidance, enabling policy optimization via RL.
   }
   \label{fig:method}

\vspace{-10pt}
\end{figure*}

\section{Method: \method}
\label{sec:method}


To learn a new task from limited demonstrations, we propose \M.
Its key insight consists of two complementary components:
(1) learning shared structure from a multi-task dataset to promote target task generalization (Section~\ref{sec:discriminator}), and
(2) acquiring recovery guidance for non-expert states through online interaction with the target environment (Section~\ref{sec:proximity}).
These components enable effective policy optimization under limited demonstrations and broad task variations (Figure \ref{fig:method}).

\subsection{Multi-task Discriminator}
\label{sec:discriminator}

To recognize expert behavior across intra-task variations,
we propose a demonstration-conditioned discriminator $d_{\phi}(\tau, (s, a))$ to facilitate knowledge transfer across tasks.
Given a task demonstration $\tau$ and a state and action pair $(s, a)$, $d_{\phi}$ predicts whether $(s, a)$ originates from the expert distribution for $\tau$'s task.
Inspired by the DVD framework \citep{chen2021learning}, we use two separate encoders to process $\tau$ and $(s, a)$, respectively. The resulting embeddings are concatenated and passed through a network composed of two fully connected layers, which outputs the probability that $(s, a)$ is consistent with expert behavior for the task described by $\tau$.

We train $d_{\phi}$ using binary classification loss on expert demonstrations from all tasks, $\mathcal{D}_{target} \cup \mathcal{D}_{multi}$.  
We sample trajectories and state-action tuples from the same task as positive classification pairs and state-action tuples from different tasks as negative pairs. 
To prevent overfitting to expert data, we additionally incorporate online policy samples by labeling them as negative pairs, analogous to adversarial imitation learning methods \citep{ho2016gail}.
The resulting loss combines binary classification between task demonstrations with the adversarial objective (Equation~\ref{eqn:dvd}), where we use $\mathcal{D}$ to denote the combined dataset $\mathcal{D}_{target} \cup \mathcal{D}_{multi}$ and $\mathcal{D}_i$ to refer to the demonstrations for task $i$ from $\mathcal{D}$.

\vspace{-10pt}
\begin{equation}
\begin{aligned}
      \label{eqn:dvd} 
  L_{multi} &= \mathbb{E}_{\tau^i, (s^i, a^i) \sim \mathcal{D}_i}[\log(1 - d(\tau^i, s^i, a^i)] \\
  & + \mathbb{E}_{\tau^i \sim \mathcal{D}_i, (s^j, a^j) \sim \mathcal{D}_j, i \neq j}[\log (d(\tau^i, s^j, a^j))] + \mathbb{E}_{\tau \sim \mathcal{D}, (s, a) \sim \pi}[\log (d(\tau, s, a))] 
\end{aligned}
\end{equation}
\vspace{-10pt}

For notational simplicity, we denote the target task discriminator as $d(s,a)$, where the demonstration $\tau$ is implicitly sampled from $\mathcal{D}_{target}$. 


\subsection{Proximity Function}
\label{sec:proximity}
To provide informative rewards in non-expert states that effectively guide the policy back towards the expert state distribution,
we introduce a proximity function $p_{\theta}(s)$.  This function estimates the temporal distance between a state $s$ and the expert state distribution (i.e., the minimum number of steps to reach an expert state).  
Directly computing this distance is impractical, as it requires rolling out the policy from each state and changes dynamically as the policy explores the environment.

Inspired by quasimetric learning~\citep{quasimetric23}, we adopt their chain analogy to build intuition for our proximity objective. Consider two objects connected by multiple chains, each consisting of links of fixed length. If the objects are pulled apart, their separation is limited by the shortest chain, revealing the length of the ``optimal'' chain. Similarly, in our setting there are multiple paths from a state $s$ to the expert distribution through policy transitions. Assuming each transition incurs a fixed temporal cost, the distance from $s$ to the expert distribution is determined by the shortest such path, which we estimate as the proximity measure.

In practice, we learn $p_\theta(s_t)$ by maximizing its value on policy states, pushing them away from the expert distribution, while enforcing local temporal consistency constraints. These constraints ensure that $p_\theta(s_t)$ is upper-bounded by the cumulative cost of any path from $s_t$ to the expert distribution. Specifically, for any transition $(s_t, s_{t+1})$, the proximity of $s_t$ can be at most one timestep greater than that of $s_{t+1}$, yielding the triangle inequality constraint $p_\theta(s_t) \leq p_\theta(s_{t+1}) + \zeta$, where $\zeta$ is a hyperparameter that defines the cost of one timestep. To ground the function, we anchor all expert states to have a proximity of $0$. States farther from the expert distribution increase by $\zeta$ per step. The overall objective is given by:



\vspace{-10pt}
\begin{equation}
\max_{\theta} \;
\underbrace{\mathbb{E}_{(s_t, s_{t+1}) \sim \mathcal{D}_\pi} p_\theta(s_t)}_{\text{maximize proximity of policy states}}
\quad \text{s.t.} \quad
\begin{cases}
\underbrace{p_\theta(s_t) \leq p_\theta(s_{t+1}) + \zeta}_{\text{temporal consistency}}, 
& \forall (s_t, s_{t+1}) \in \mathcal{D}_\pi \\
\underbrace{p_\theta(s_{e}) = 0}_{\text{expert boundary condition}}, 
& \forall s_{e} \in \mathcal{D}_{target}
\end{cases}
\end{equation}
\vspace{-10pt}

We optimize this objective using a Lagrangian relaxation, yielding the following maximization problem, where $(x)^+ = \max(x,0)$:

\vspace{-10pt}
\begin{equation}
\label{eqn:prox}
\max_{\theta} \;
\mathbb{E}_{(s_t, s_{t+1}) \sim \mathcal{D}_\pi} \Big[
\underbrace{p_\theta(s_t)}_{\text{maximize  proximity}}
-
\underbrace{\alpha \left(p_\theta(s_t) - p_\theta(s_{t+1}) - \zeta\right)^+}_{\text{temporal consistency penalty}}
\Big]
-
\underbrace{\beta \, \mathbb{E}_{s_{e} \sim \mathcal{D}_{target}} |p_\theta(s_{e})|}_{\text{expert boundary penalty}}
\end{equation}
\vspace{-10pt}

We set fixed Lagrangian multipliers ($\alpha = 100$, $\beta = 5$) and bound the output of $p_\theta$ to the range $[0,1]$ using a sigmoid activation. While the discriminator operates in a multi-task setting, $p_\theta$ is trained \emph{only} on target-task demonstrations and policy samples. To leverage information from the multi-task dataset and enrich the expert distribution for the target task, we additionally re-label policy states as expert if the multi-task discriminator's confidence exceeds a threshold, i.e., $d(s_t, a_t) > c_{thresh}$.

\subsection{\method}
Our full reward function $\tilde{R}$, shown in Equation~\ref{eqn:R_tilde}, combines the sparse, generalizable signal from the discriminator with the dense, informative guidance from the proximity function. 
Specifically, the discriminator score $d(s,a)$ is used directly as a reward to encourage expert-consistent behavior. 
The agent should also be rewarded for transitions that \emph{reduce} its proximity to the expert distribution, so we incorporate the proximity improvement $p(s_t) - p(s_{t+1})$ with weight $\lambda$. 

\vspace{-0pt}
\begin{equation}
\label{eqn:R_tilde}
     \tilde{R}(s_{t},a_{t},s_{t+1}) = \underbrace{d(s_{t},a_{t})}_{\text{expert behavior signal}} + \lambda \underbrace{\big[p(s_{t}) - p(s_{t+1})\big]}_{\text{proximity improvement}}
\end{equation}

With the reward function defined, we summarize the training procedure of \M~in Algorithm~\ref{alg:method}.
At each iteration, the policy collects transitions from the environment, after which we update $d_{\phi}$ and  $p_{\theta}$. The policy $\pi$ is then optimized using RL with rewards from $\tilde{R}$. 
Implementation details and hyperparameters are provided in Appendix~\ref{sec:mpirl_implementation} and Appendix~\ref{sec:mpirl_hyperparameters}. 

\begin{algorithm}[ht]
\small
\caption{\M}
\label{alg:method}
\begin{algorithmic}[1]
\State \textbf{Input:} Target-task demos $\mathcal{D}_{target}$, multi-task demos $\mathcal{D}_{multi}$
\State Initialize policy $\pi$, discriminator $d_{\phi}$, proximity function $p_{\theta}$, replay buffer $\mathcal{D}_\pi$

\For{$i = 1,2,\dots,N$}
    \State Collect transitions
    $(s_t, a_t, s_{t+1})_{t=0}^{T}$ by rolling out $\pi$ in the target-task environment
    
    \State Add collected transitions to replay buffer  $\mathcal{D}_\pi$

    \State Update discriminator $d_{\phi}$ using Eq.~\ref{eqn:dvd} 
    
    \State Update proximity function $p_{\theta}$ using Eq.~\ref{eqn:prox} 
     
    \State Update policy $\pi$ via RL with reward $\tilde{R}$ from Eq.~\ref{eqn:R_tilde}
\EndFor

\State \textbf{Output:} Trained policy $\pi$
\end{algorithmic}
\end{algorithm}
\vspace{-10pt}

\begin{figure*}[t]
    \centering
    \begin{subfigure}[b]{0.22\textwidth}
    \centering
    \includegraphics[width=0.9\textwidth]{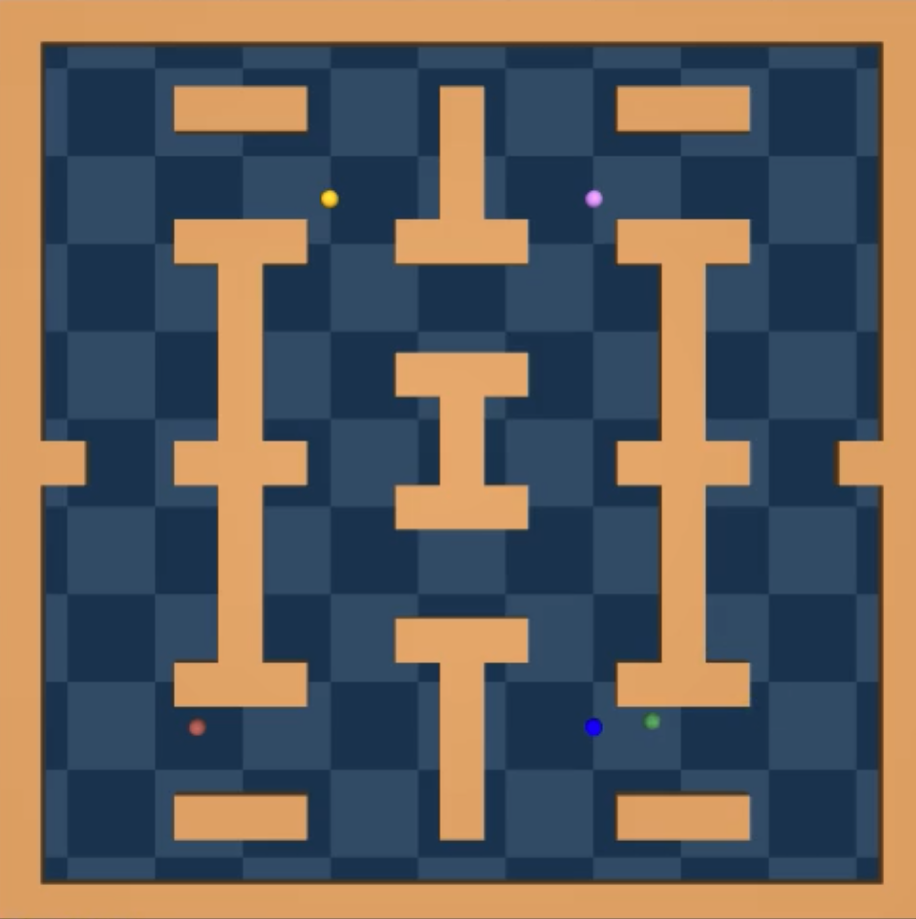}
    \caption{Maze2D}
    \label{fig:env_maze}
    \end{subfigure}
    \hfill
    \begin{subfigure}[b]{0.22\textwidth}
    \centering
    \includegraphics[width=0.9\textwidth]{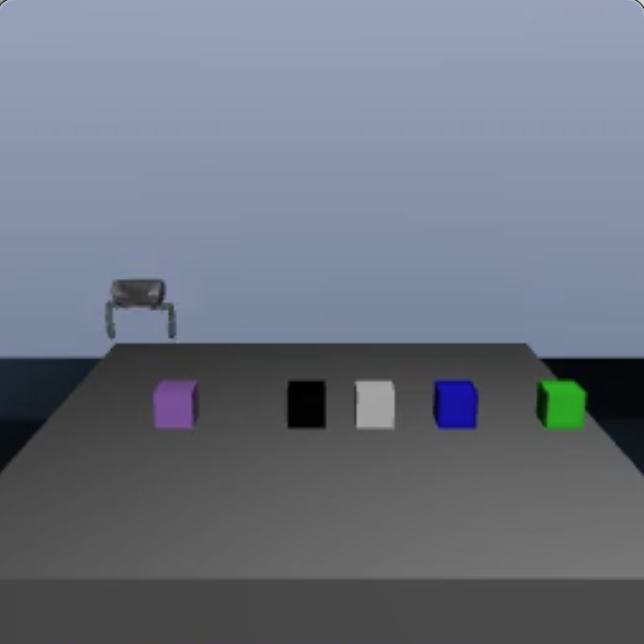}
    \caption{Block Stacking}
    \label{fig:env_block_stacking}
    \end{subfigure}
    \hfill
    \begin{subfigure}[b]{0.45\textwidth}
    \centering
    \includegraphics[width=0.9\textwidth]{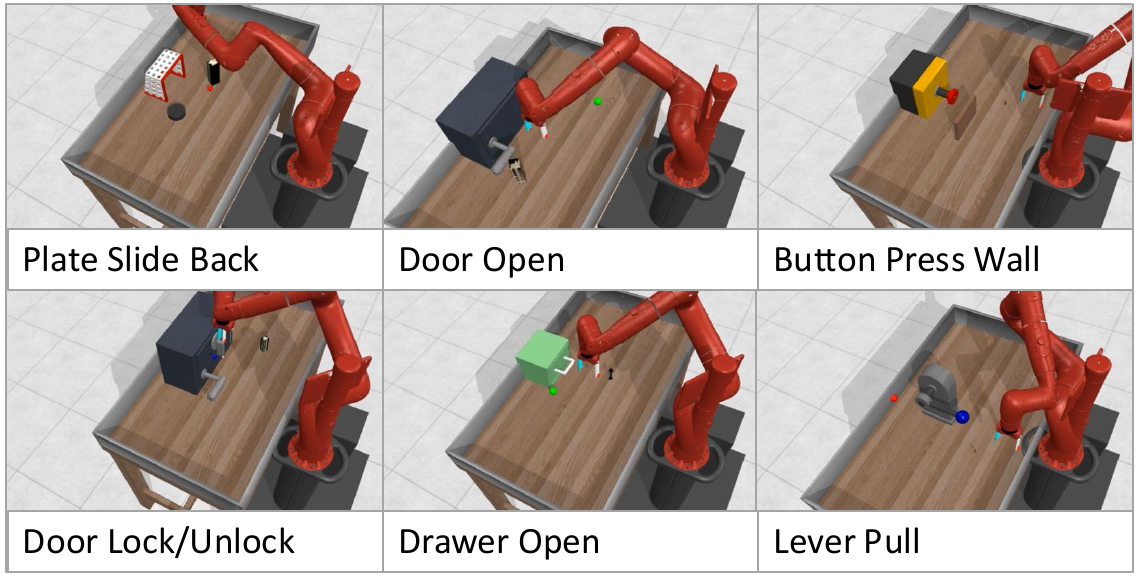}
    \caption{FactorWorld}
    \label{fig:env_factorworld}
    \end{subfigure}
    \caption[]{
    We evaluate on navigation and manipulation tasks with diverse task configurations.
    }
    \label{fig:environment}
    \vspace{-10pt}
\end{figure*}

\section{Experiments}
\label{sec:results}

We answer the following questions in our experiments: (1) How effective is \M~compared to other methods that learn from limited demonstrations? 
(2) How does \M's performance vary with the number of demonstrations and tasks in the multi-task dataset? 
(3) How do the components of \M~contribute to its performance?

\textbf{Environment.} We evaluate our methods on several IRL tasks in different environments (Figure~\ref{fig:environment}):
\emph{Maze2D} from the D4RL benchmark \citep{fu2020d4rl}: the agent navigates to a fixed goal location, distinguished by color, among four fixed objects. 
The intra-task variation comes from the agent’s random starting position.  
\emph{Block Stacking}  \citep{pertsch2021accelerating}: the agent’s goal is to pick up a block of color X and place it on a block of color Y. Differently colored blocks have random initial positions creating intra-task variation. 
\emph{FactorWorld} from \citet{xie2024decomposing}: a multi-task benchmark of manipulation tasks with intra-task variations in object position, table position, distractor objects \& positions, and arm position. Each task additionally exhibits distinct dynamics.
Further details for all environments are provided in Appendix~\ref{sec:appendix_environment}.

\textbf{Baselines.}
To the best of our knowledge, no prior work directly tackles the \mysetting~setting.
So we compare with state-of-the-art methods from closely related settings, augmenting them with additional assumptions where possible for a fair comparison (details in Appendix \ref{sec:appendix_implementation_details}). In short, we consider:
\emph{BC} clones the target task demonstrations only.
We further include two BC variants: \emph{DemoConditioned-BC}, which pretrains a demonstration-conditioned policy on $\mathcal{D}_{multi}$ and fine-tunes it on $\mathcal{D}_{target} \cup \mathcal{D}_{multi}$; and \emph{Transformer-BC}, a Transformer-based policy from \citet{robomimic2021}, trained with the same pretraining and fine-tuning protocol.
\emph{SQIL}~\citep{reddy2020sqil} is state-of-the-art online IL method.
\emph{GAIL}~\citep{ho2016gail} is a widely used IRL method.
\emph{MT-AIRL} is AIRL~\citep{fu2018airl}, an adversarial maximum entropy IRL method trained on multi-task demonstrations.
To provide a comprehensive comparison, we  include a meta-IRL baseline,  \emph{PEMIRL}~\citep{yu2019meta}, which meta-learns a reward function with access to multi-task training environments. 
\emph{GoalPro}~\citep{lee2021generalizable} learns a goal-proximity reward. 
\emph{DVD}~\citep{chen2021learning} learns a multi-task discriminator only using the demonstrations; we evaluate the discriminator with online RL.


\begin{table*}[t]
    \centering
    \caption{
   Comparison of mean policy success rates (\%) under the \mysetting~setting across nine navigation and manipulation tasks. We compare \M~against nine baselines and report 95\% confidence intervals over five random seeds. The best and second-best mean performances on each task are shown in \textbf{bold} and \underline{underlined}, respectively.
    }
    \label{tab:main_results}
    \setlength{\tabcolsep}{2.5pt}
    \renewcommand{\arraystretch}{1.08}
    \scriptsize
    \resizebox{\textwidth}{!}{%
    \begin{tabular}{lccccccccc}
        \toprule
        Method
        & Maze2D
        & \shortstack{Block\\Stacking}
        & \shortstack{Drawer\\Open}
        & \shortstack{Door\\Lock}
        & \shortstack{Door\\Unlock}
        & \shortstack{Plate Slide\\Back}
        & \shortstack{Door\\Open}
        & \shortstack{Lever\\Pull}
        & \shortstack{Button Press\\Wall} \\
        \midrule

        GAIL
        & $53.2 \pm 37.7$
        & $\underline{25.9 \pm 43.9}$
        & $17.7 \pm 10.9$
        & $22.0 \pm 6.3$
        & $43.9 \pm 17.3$
        & $\underline{35.5 \pm 21.4}$
        & $\underline{50.2 \pm 34.1}$
        & $\underline{52.2 \pm 45.1}$
        & $22.2 \pm 9.2$ \\

        MT-AIRL
        & $42.9 \pm 12.7$
        & $9.1 \pm 16.1$
        & $25.3 \pm 4.6$
        & $33.0 \pm 10.8$
        & $32.6 \pm 17.4$
        & $19.3 \pm 13.2$
        & $26.4 \pm 12.5$
        & $22.7 \pm 19.4$
        & $23.4 \pm 11.7$ \\

        PEMIRL
        & $15.4 \pm 8.4$
        & $0.7 \pm 2.0$
        & $34.2 \pm 11.7$
        & $15.4 \pm 5.6$
        & $7.1 \pm 3.4$
        & $13.3 \pm 1.5$
        & $4.5 \pm 6.3$
        & $5.9 \pm 3.8$
        & $2.5 \pm 3.5$ \\

        \midrule

        BC
        & $53.6 \pm 16.2$
        & $22.0 \pm 19.8$
        & $30.5 \pm 1.6$
        & $37.3 \pm 16.0$
        & $28.7 \pm 25.0$
        & $30.5 \pm 8.4$
        & $49.3 \pm 23.5$
        & $44.5 \pm 21.3$
        & $38.0 \pm 5.0$ \\

        DemoConditioned-BC
        & $15.4 \pm 11.1$
        & $0.4 \pm 0.7$
        & $57.2 \pm 21.6$
        & $\underline{52.0 \pm 16.7}$
        & $\underline{61.4 \pm 20.1}$
        & $28.9 \pm 4.9$
        & $47.4 \pm 24.8$
        & $34.7 \pm 22.0$
        & $\underline{69.0 \pm 25.6}$ \\

        Transformer-BC
        & $21.5 \pm 7.5$
        & $5.1 \pm 9.2$
        & $\underline{66.4 \pm 23.0}$
        & $35.9 \pm 14.9$
        & $44.6 \pm 16.3$
        & $33.7 \pm 8.3$
        & $25.7 \pm 38.3$
        & $8.6 \pm 4.7$
        & $51.3 \pm 24.9$ \\

        SQIL
        & $51.5 \pm 5.4$
        & $6.0 \pm 11.4$
        & $10.8 \pm 12.1$
        & $15.8 \pm 2.5$
        & $6.0 \pm 0.0$
        & $14.6 \pm 8.7$
        & $3.7 \pm 0.6$
        & $14.8 \pm 11.3$
        & $6.4 \pm 4.2$ \\

        \midrule

        DVD
        & $31.7 \pm 33.5$
        & $0.0 \pm 0.0$
        & $9.0 \pm 8.9$
        & $24.8 \pm 4.1$
        & $14.9 \pm 15.6$
        & $12.8 \pm 5.0$
        & $13.5 \pm 5.9$
        & $12.2 \pm 11.2$
        & $12.2 \pm 6.3$ \\

        Goal Proximity
        & $\mathbf{96.0 \pm 2.8}$
        & $1.0 \pm 2.0$
        & $7.4 \pm 7.3$
        & $14.8 \pm 5.6$
        & $7.2 \pm 6.8$
        & $27.6 \pm 5.1$
        & $4.6 \pm 9.0$
        & $0.0 \pm 0.0$
        & $12.3 \pm 22.5$ \\

        \midrule

        Ours
        & $\underline{94.1 \pm 3.6}$
        & $\mathbf{61.6 \pm 19.9}$
        & $\mathbf{94.1 \pm 2.2}$
        & $\mathbf{57.6 \pm 31.8}$
        & $\mathbf{78.6 \pm 22.0}$
        & $\mathbf{63.9 \pm 23.7}$
        & $\mathbf{97.0 \pm 1.1}$
        & $\mathbf{93.2 \pm 3.0}$
        & $\mathbf{90.5 \pm 5.8}$ \\

        \bottomrule
    \end{tabular}%
    }
    \vspace{-10pt}
\end{table*}

\subsection{Quantitative Results under \mysetting}
\label{sec:main_comparison}

As shown in Table~\ref{tab:main_results}, \M~achieves the highest mean success rate on eight of the nine tasks. Across all tasks, \M~improves over the strongest per-task baseline by an average of \textbf{24.7 percentage points} in success rate. More  details can be found in Appendix~\ref{sec:appendix_more_main_results}. 



\textbf{\M~vs. IRL Baselines.}
We begin by comparing \M~against IRL methods. Although GAIL has access to the same data resources as \M~(few-shot target demonstrations, multi-task demonstrations, and policy samples), its performance remains poor in our setting, achieving only 35.9\% average success. This suggests that GAIL cannot effectively leverage these heterogeneous sources of information. 
Similarly, MT-AIRL reaches only 26.1\% average success, indicating limited robustness to scarce demonstrations and task variation.
PEMIRL performs even worse, with 11.0\% average success, due to the high sample cost of meta-training within a fixed training budget, highlighting the inefficiency of meta-IRL when targeting a single task. 

\textbf{\M~vs. IL Baselines.}
We next compare against IL baselines. BC and its variants are relatively strong among the baselines, with DemoConditioned-BC achieving the best baseline average success rate of 40.7\%. This suggests  that under limited demonstrations and substantial variations, inferring a reliable reward function is more challenging than directly learning a policy.
In contrast, SQIL learns an RL policy but relies on an extremely sparse reward signal, which leads to poor performance on most tasks, with 14.4\% average success. The main exception is Maze2D, where the environment configuration is fixed and demonstrated trajectories can be more easily revisited.

\textbf{\M~vs. Discriminator/Proximity Baselines.}
Finally, we compare with prior work that shares a similar objective with parts of our method.
GoalPro learns a proximity-based reward along  \emph{expert trajectories} rather than in \emph{non-expert states}. While it performs well on Maze2D, achieving 96.0\% success due to the fixed goal and environment configuration, it fails on most manipulation tasks and obtains only 19.0\% average success. 
Despite using a multi-task success discriminator, DVD performs poorly, with 14.6\% average success, due to its pre-trained reward function, which can be exploited by the policy. 


\subsection{Analysis: Sensitivity to Demonstration Quantity and Task Diversity}
\label{sec:analysis}

\begin{wrapfigure}{r}{0.58\linewidth}
    \centering
    \vspace{-1.2em}
    \begin{subfigure}[t]{0.48\linewidth}
        \centering
        \includegraphics[width=\linewidth]{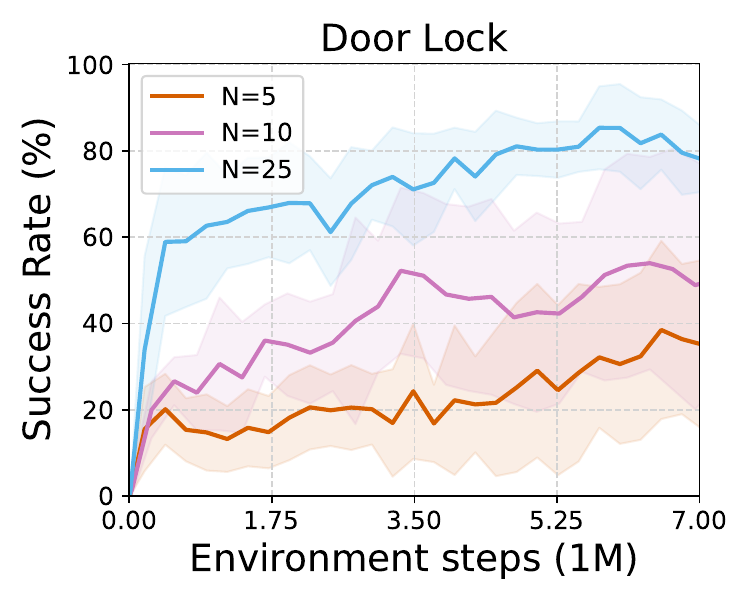}
        \caption{Number of Target Demos}
        \label{fig:analysis:demo_num}

    \end{subfigure}
    \hfill
    \begin{subfigure}[t]{0.48\linewidth}
        \centering
        \includegraphics[width=\linewidth]{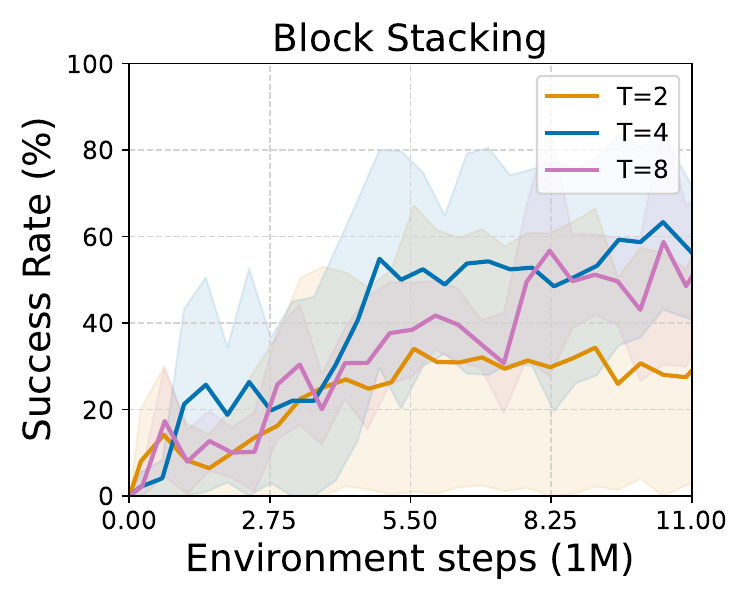}
        
        \caption{Number of Tasks in $\mathcal{D}_{multi}$}
        \label{fig:analysis:mt_num}
    \end{subfigure}
    \caption{Data efficiency analysis.
    Effect of (a) demonstration quantity and (b) task diversity.
    }

    \label{fig:sensitivity_analysis}
    \vspace{-20pt}
\end{wrapfigure}

\textbf{Demonstration Quantity.} 
We analyze the effect of demonstration quantity on the Door-Lock task. As shown in Figure~\ref{fig:analysis:demo_num}, performance improves as the number of target demonstrations increases from $5$ to $25$. This suggests that additional demonstrations better characterize the expert distribution, enabling improved policy learning. Results on additional tasks are provided in Appendix~\ref{sec:appendix_analysis}.
To further understand performance when demonstrations are abundant, we also evaluate a setting with 200 demonstrations. 
In this regime, \M~continues to perform strongly, achieving an average success rate of 85\% across five tasks, while representative baselines—BC, SQIL, and GAIL—reach 70\% on average. 
This indicates that \M~performs well both with limited and ample demonstrations, whereas the baselines require abundant demonstrations to achieve strong performance (see Appendix~\ref{sec:appendix_200} for details).

\textbf{Task Diversity.} 
We analyze the effect of task diversity by varying the number of tasks $T$ in the multi-task dataset for the Block-Stacking task. As shown in Figure~\ref{fig:analysis:mt_num}, performance improves when $T$ increases from $2$ to $4$, but plateaus from $4$ to $8$. This suggests that once a moderate level of task diversity is reached, additional tasks contribute limited new information. Analysis on more tasks can be found in Appendix~\ref{sec:appendix_analysis}.

\subsection{Ablations on \M}
\label{sec:ablations}
\begin{figure*}[ht]
    \centering
    \begin{subfigure}[t]{0.3\linewidth}
        \centering
        \raisebox{0\height}{\includegraphics[width=\linewidth]{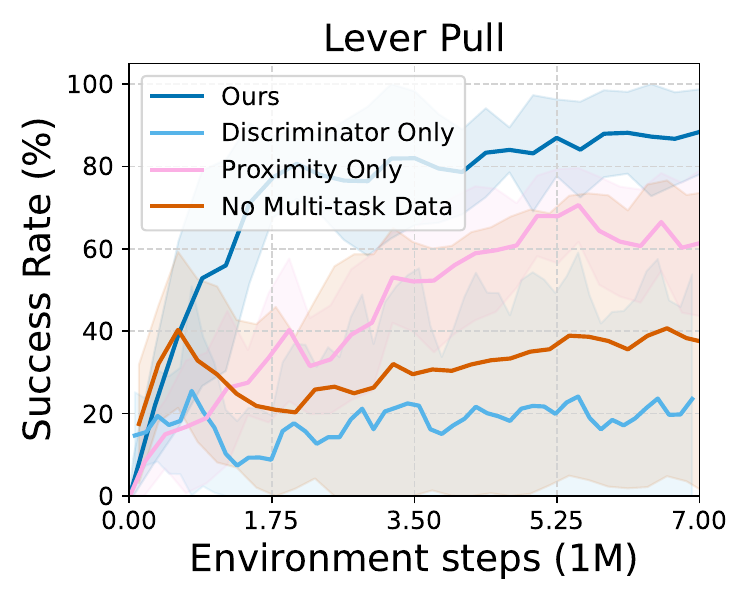}}
         \vspace{-15pt}
        \caption{Ablating Reward Terms}
        \label{fig:ablation_reward}
    \end{subfigure}
    \begin{subfigure}[t]{0.29\linewidth}
        \centering
        \raisebox{0\height}{\includegraphics[width=\linewidth]{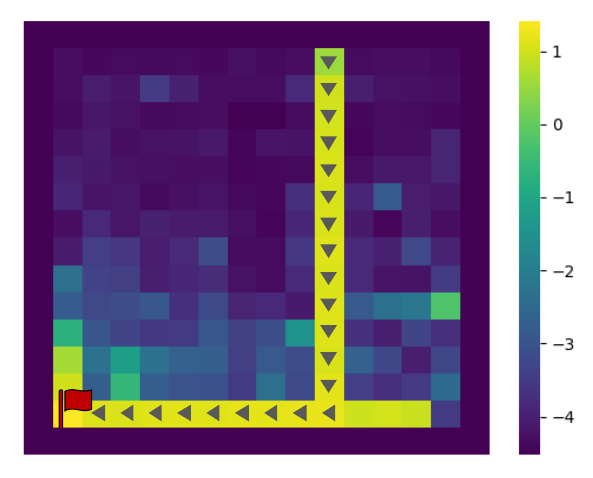}}
         \vspace{-15pt}
        \caption{Discriminator Heatmap}
        \label{fig:heatmap_discrimiantor}
    \end{subfigure}
    \begin{subfigure}[t]{0.28\linewidth}
        \centering
        \raisebox{0\height}{\includegraphics[width=\linewidth]{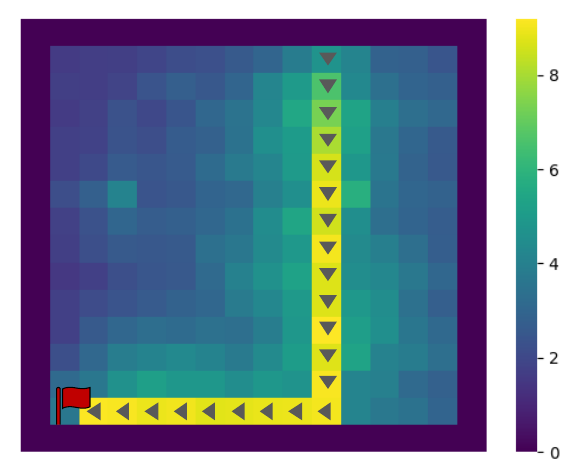}}
         \vspace{-15pt}
        \caption{Proximity Heatmap}
        \label{fig:heatmap_prox}
    \end{subfigure}
    \caption[]{Ablation of \M’s reward components. 
    }
    \label{fig:ablation}
    \vspace{-10pt}
\end{figure*}

We first ablate the two parts of our reward function $\tilde{R}$ by training policies with either a \textsc{Discriminator Only} reward or a \textsc{Proximity Only} reward.  
In Figure~\ref{fig:ablation_reward}, while each part individually provides benefits, combining both yields the best performance for \M. Additionally,  removing the multi-task dataset (\textsc{No Multi-task Data}) degrades performance, indicating that information from the dataset is important. See Appendix~\ref{sec:appendix_ablation} for additional results and details.  

We further illustrate the two components in a simple empty Minigrid environment \citep{MinigridMiniworld23} (Figure \ref{fig:heatmap_discrimiantor}, \ref{fig:heatmap_prox}). The red flag indicates the goal and arrows show the expert demonstration. Lighter colors correspond to higher rewards. Appendix~\ref{env:minigrid} gives more details about this environment.
The two parts of our reward function provide complementary and informative learning signal: the discriminator generalizes to  certain goal-reaching paths (e.g., directly above the goal), while the proximity provides a smooth gradient in non-expert regions. 

Then, we evaluate the standalone contribution of the proximity reward by augmenting GAIL with our proximity function. In 3 out of 5 tasks, this improves GAIL's performance and stability, highlighting the usefulness of the proximity reward for providing guidance in non-expert regions. See details and results in Appendix~\ref{sec:appendix_prox}. Finally, we evaluate \M~under different hyperparameter setting and observe that it remains robust without careful fine-tuning (see Appendix~\ref{sec:appendix_ablation} for details).



\section{Conclusion}
\label{sec:conclusion}
We introduce a novel problem setting: few-shot IRL with multi-task demonstrations, aiming to learn a new task with diverse variations.  To solve it, we propose \M, a novel method that learns a two-part reward function: (1) a multi-task discriminator that recognizes expert behavior over task variations, and (2) a proximity reward that provides guidance in non-expert states.  Finally, we demonstrate the effectiveness of our method across multiple navigation and manipulation environments, where it achieves the highest mean success rate on eight of the nine tasks and improves over the strongest per-task baseline by 24.7 percentage points on average. We provide further discussion in Appendix~\ref{sec:appendix_limitation}.


\bibliography{main}
\bibliographystyle{rlj}

\beginSupplementaryMaterials

\appendix



\section {Code}
\label{sec:appendix_algorithm}

We have made our code and data available to download here \url{https://drive.google.com/drive/folders/1JbN2GX0005qrPSvRD3IEASOa9o5S3SMs?usp=sharing}.






        






\section{Additional Results}
\label{sec:appendix_results}
\subsection{Additional Details for the Main Results Table}
\label{sec:appendix_more_main_results}

Table~\ref{tab:main_results} reports the performance comparison on Maze2D, Block Stacking, and seven FactorWorld manipulation tasks.
For each method, we report the mean success rate and 95\% confidence intervals over 5 random seeds, with 10 rollout episodes per evaluation.
For GAIL, we evaluate two variants in which the multi-task dataset is treated as either negative or positive examples when training the discriminator, but neither variant achieves competitive performance. Here we only show the result of GAIL with multi-task dataset as negative examples. In Maze2D, SQIL achieves over 50\%  success, because variations of this task arise primarily from the agent’s trajectory, where the environment configuration is fixed and the demonstrated trajectories can be revisited. 

\begin{figure*}[ht]
    \vspace{-10pt}
    \centering
    \includegraphics[width=0.325\linewidth]{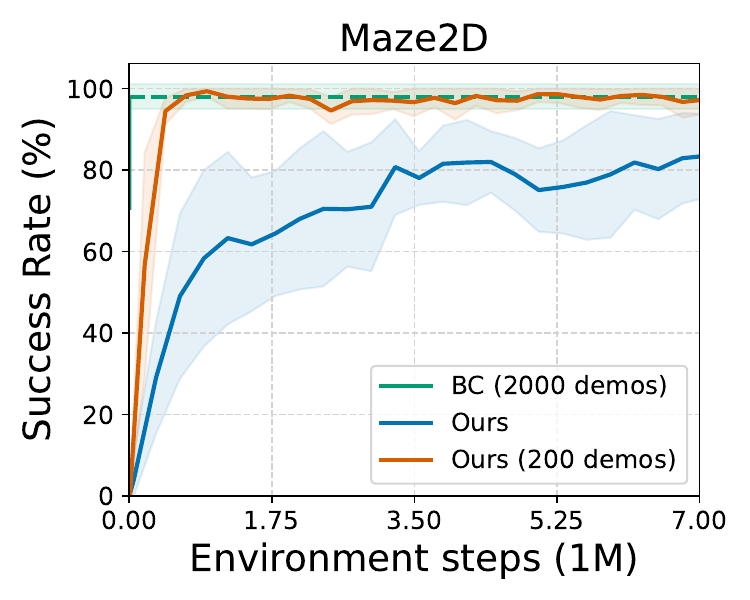}
    \includegraphics[width=0.325\linewidth]{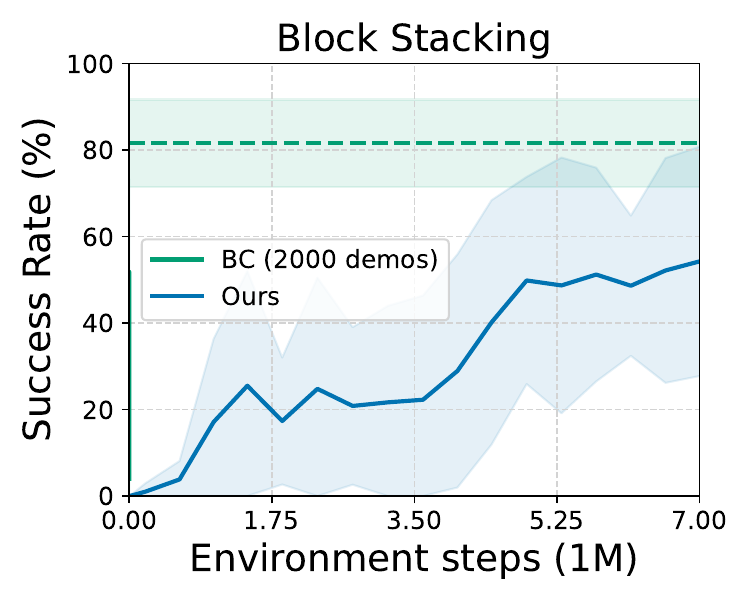}
    \includegraphics[width=0.325\linewidth]{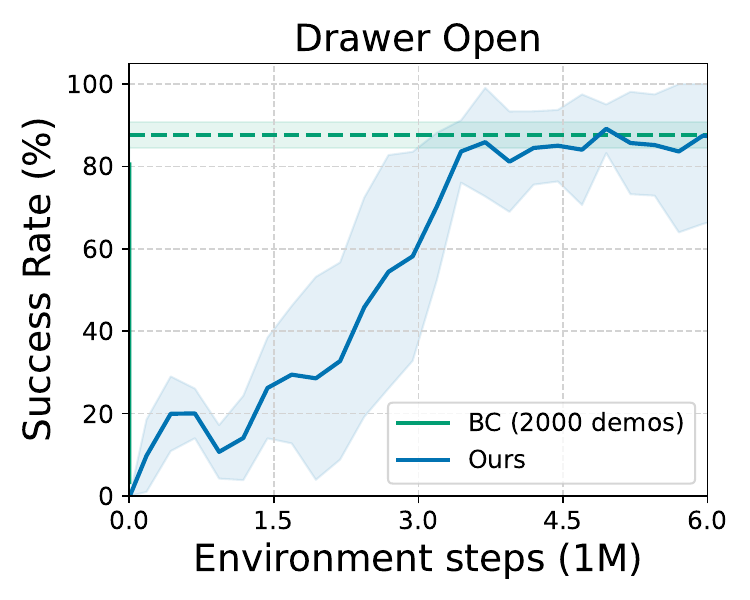}

    \includegraphics[width=0.325\linewidth]{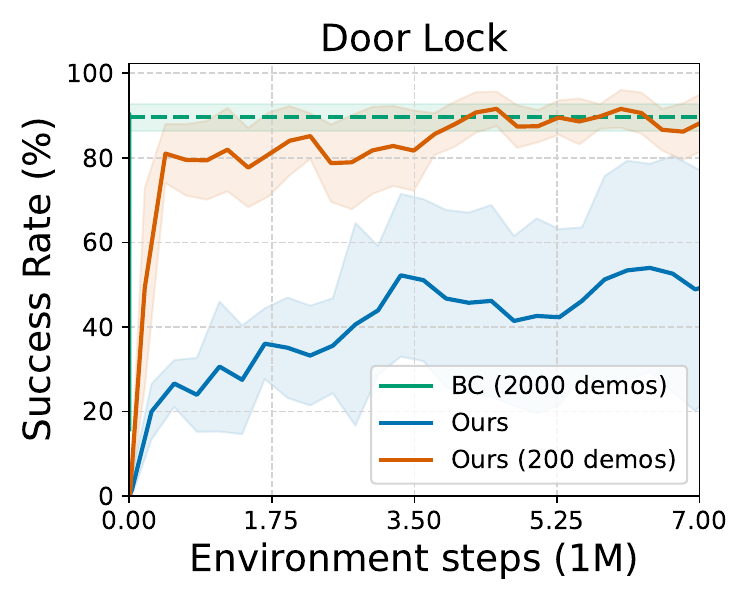}
    \includegraphics[width=0.325\linewidth]{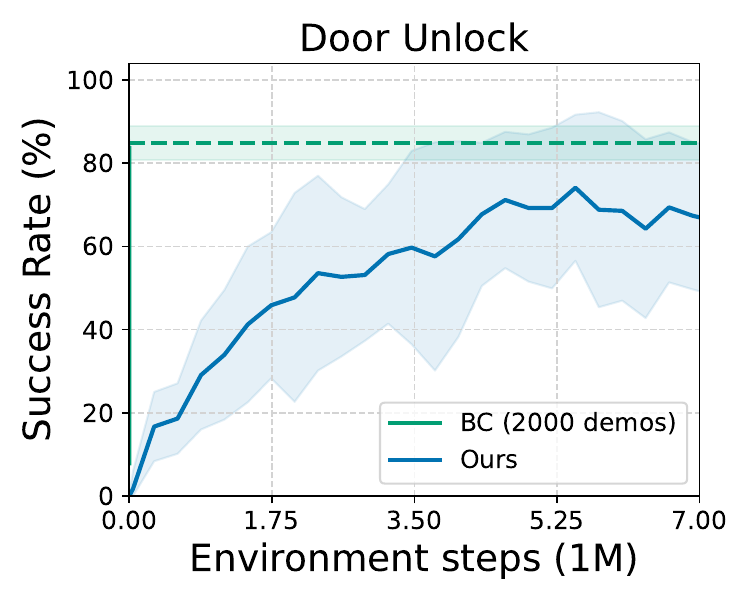}
    \includegraphics[width=0.325\linewidth]{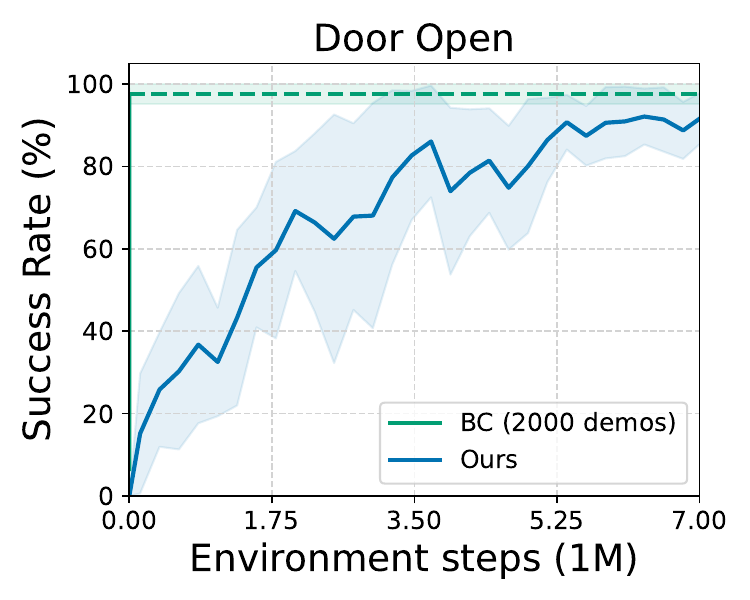}    

    \includegraphics[width=0.325\linewidth]{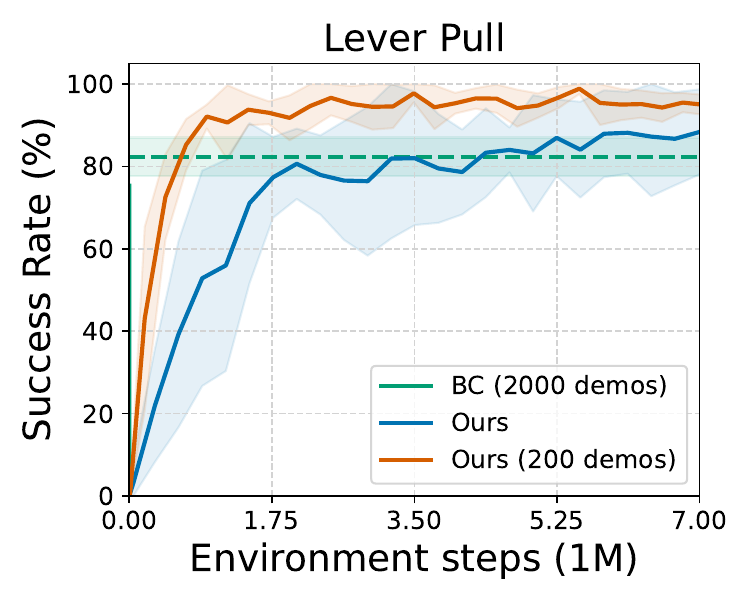}
    \includegraphics[width=0.325\linewidth]{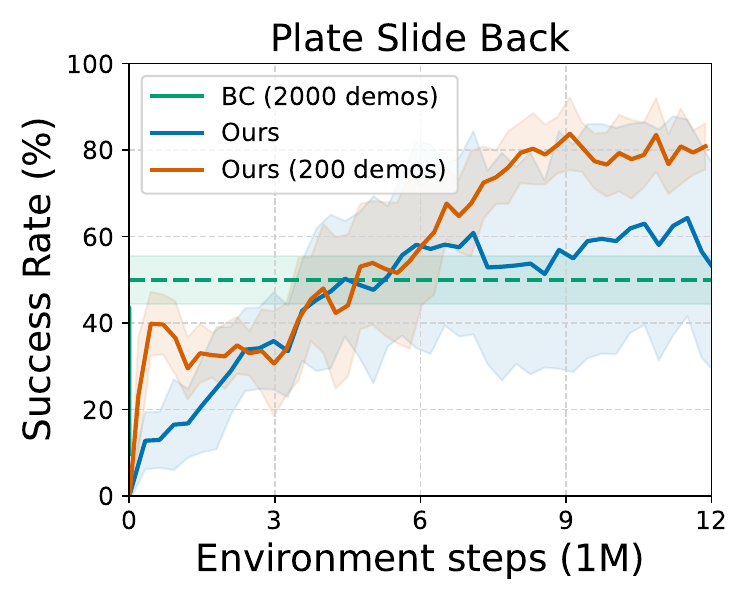}
    \includegraphics[width=0.325\linewidth]{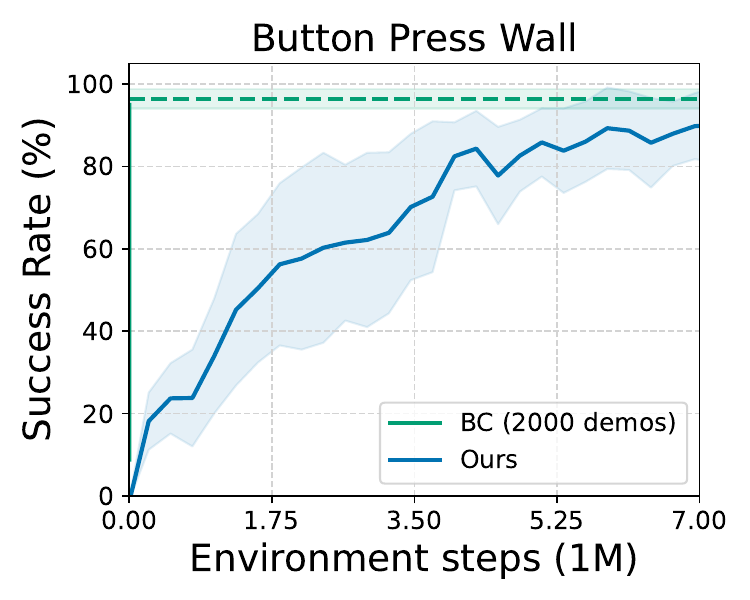}      
    \caption[]{Comparisons with an ``oracle'' BC method given 2000 demonstrations.
    }
    \label{fig:appendix_oracle_bc}

\end{figure*}

Figure~\ref{fig:appendix_oracle_bc} compares our method to an ``oracle'' BC policy trained with 2000 target-task demonstrations, serving as an approximate upper bound on imitation performance. 
This oracle is imperfect, as BC remains susceptible to compounding errors and covariate shift. 
Moreover, many of these tasks are very difficult RL problems (e.g., long-horizon maze navigation with narrow corridors, block stacking from random initial states), where training RL from scratch without demonstrations fails to reach oracle-level performance.



\begin{figure*}[t]
    \centering
    \includegraphics[width=0.325\linewidth]{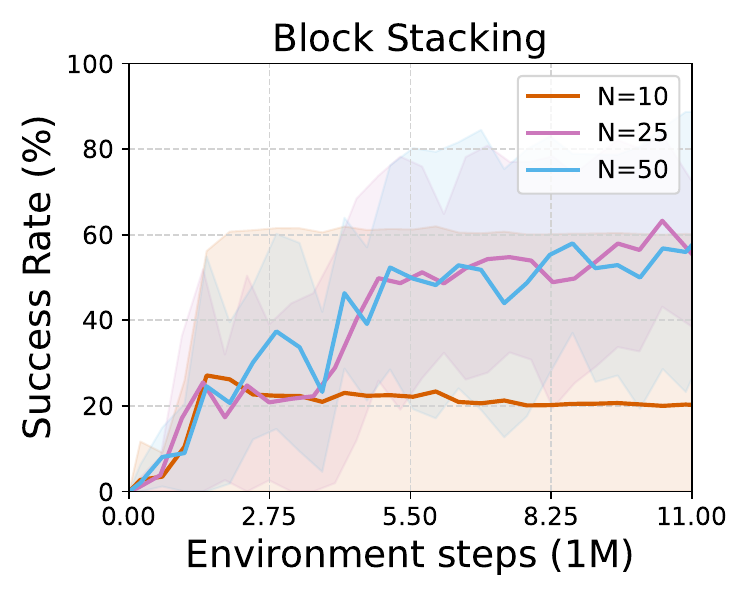}
    \includegraphics[width=0.325\linewidth]{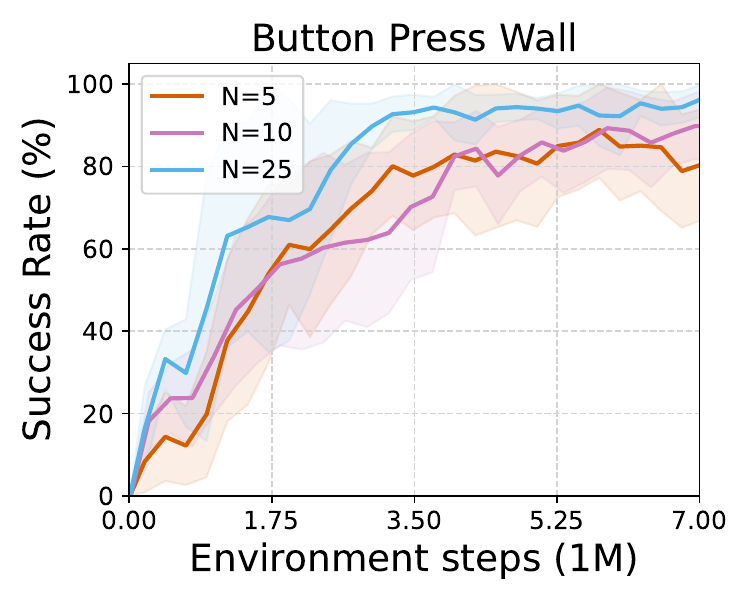}
    \includegraphics[width=0.325\linewidth]{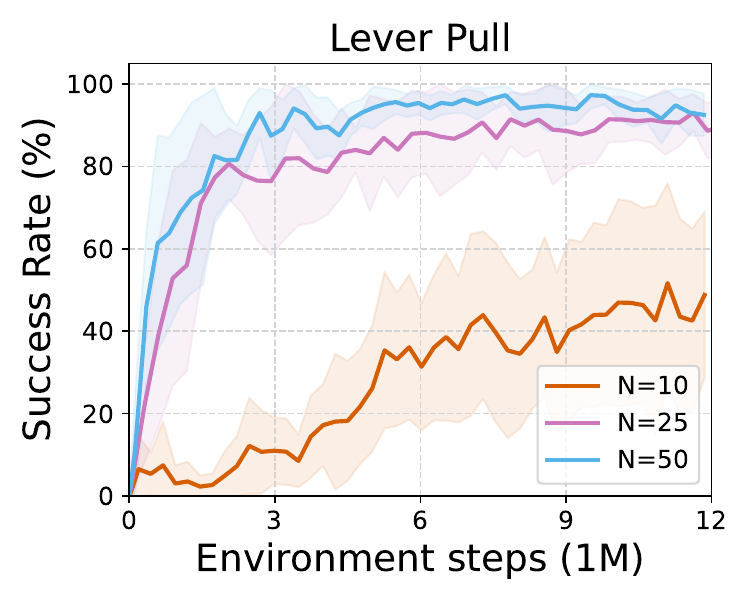}
    \includegraphics[width=0.325\linewidth]{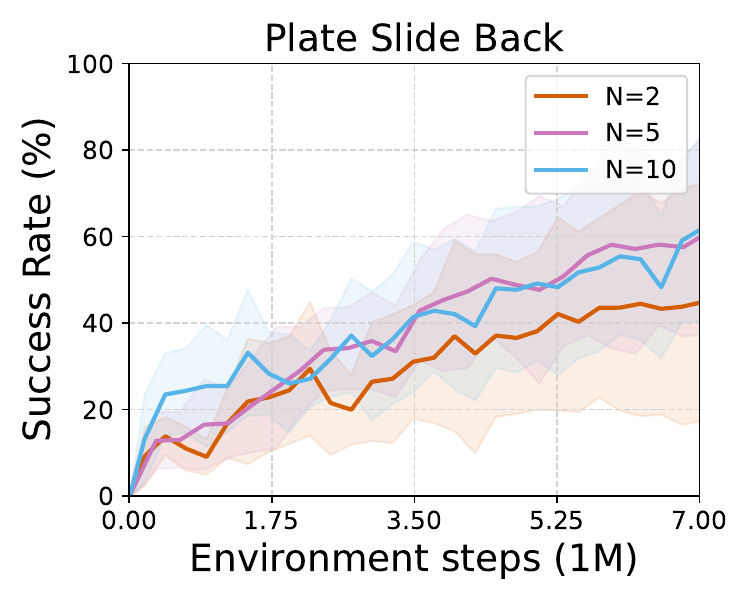}
    \includegraphics[width=0.325\linewidth]{figures/results/demo_num/door_lock_demo_num_title_success_legend.pdf}
    
    \caption[]{Analysis on the number of target task demonstrations in more tasks.
    }
    \label{fig:appendix_num_demos}
    \vspace{-10pt}
\end{figure*}

\begin{figure*}[t]
    \centering
    \includegraphics[width=0.325\linewidth]{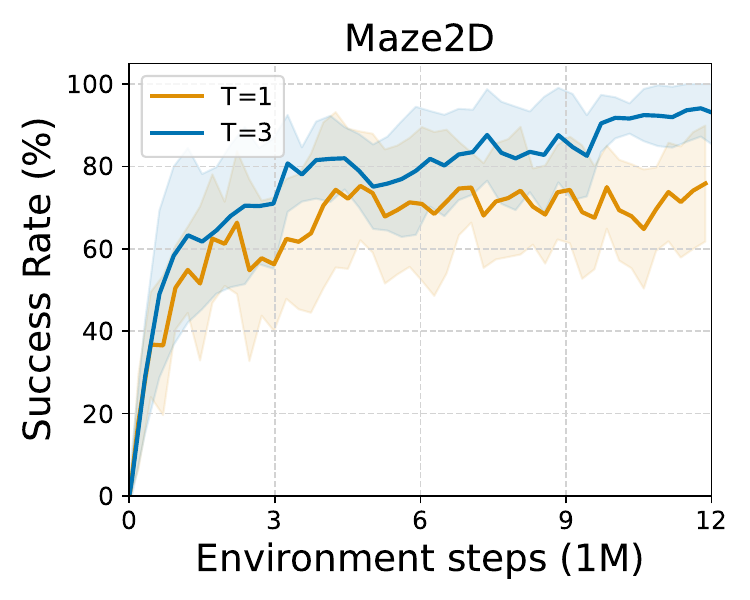}
    \includegraphics[width=0.325\linewidth]{figures/results/mt_num/block_num_tasks_success_legend.pdf}
    \includegraphics[width=0.325\linewidth]{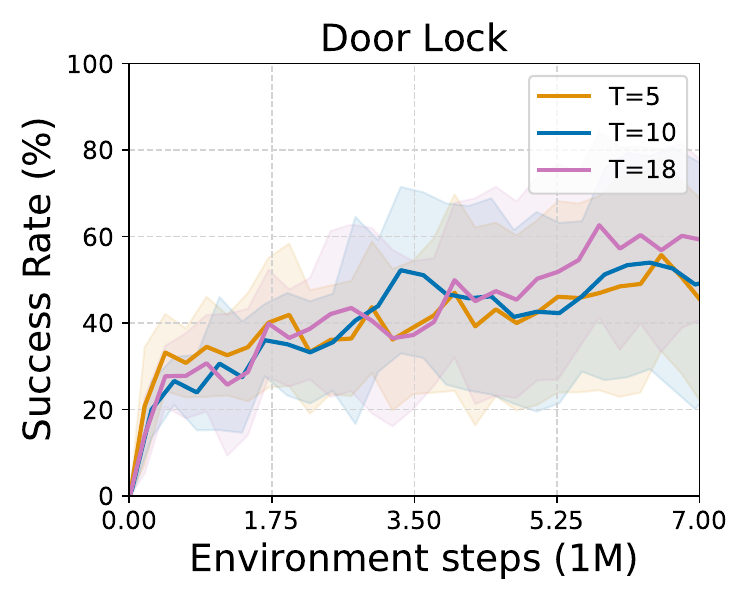}
    \includegraphics[width=0.325\linewidth]{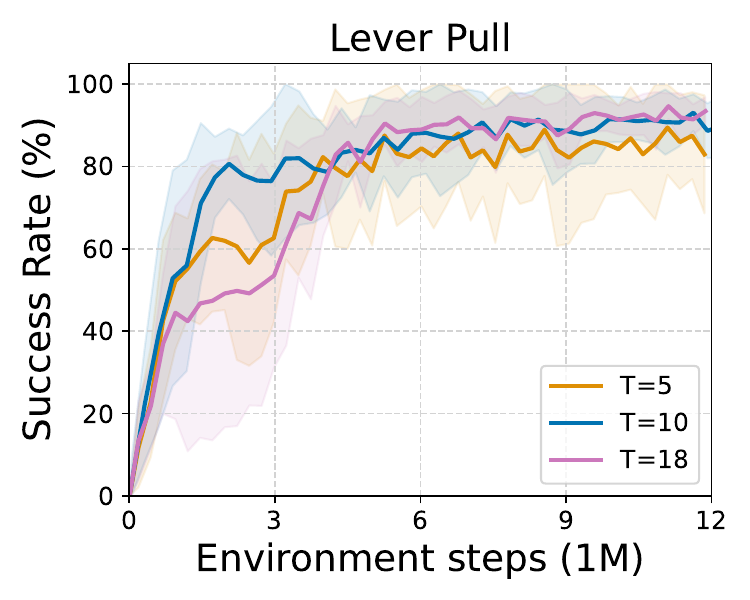}
    \includegraphics[width=0.325\linewidth]{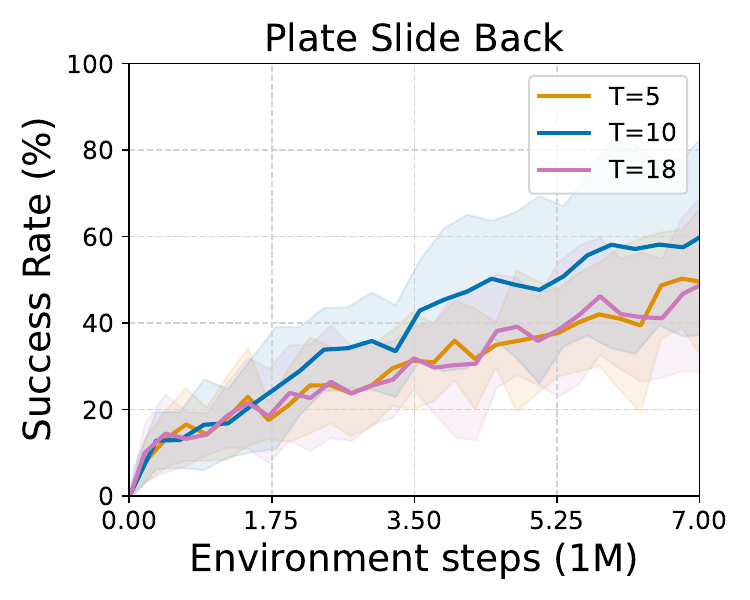}    
    \caption[]{Analysis on the number of tasks in the multi-task dataset in more tasks. 
    }
    \label{fig:appendix_num_tasks}
    \vspace{-10pt}
\end{figure*}

\subsection{Sensitivity to Demonstration Quantity and Task Diversity on More Tasks}
\label{sec:appendix_analysis}

For analysis and ablations, we evaluate on Maze, Block Stacking, and 3 out of 7 representative FactorWorld tasks.
Figure~\ref{fig:appendix_num_demos} and Figure~\ref{fig:appendix_num_tasks} report additional experiments varying  (i) the number of target task demonstrations and (ii) the number of tasks in the multi-task demonstration  dataset. We plot the average and standard deviation (in shaded regions) over 5 seeds per experimental condition and roll out 10 episodes per evaluation.

\textbf{Demonstration Quantity.}
Overall we see a similar trend as discussed in Section ~\ref{sec:analysis}, where performance increases with the number of target demonstrations until some saturation level. For example, in Button Press Wall, \M~is able to achieve 80\% success with only $5$ demonstrations.  For tasks that saturate earlier, such as Block Stacking and Plate Slide Back, it is probably that \M~saturates at this level and the remaining 40\% performance requires a more sophisticated reward function or RL algorithm.  

\textbf{Task Diversity.}
Increasing the number of tasks $T$ in the multi-task dataset sometimes yields negligible improvement (e.g., Lever-Pull), whereas in other tasks (e.g., Block-Stacking), performance improves when increasing $T$ from $2$ to $4$. 
This suggests that once a minimum level of task diversity is reached, performance saturates.   

\subsection{Performance with Ample Target-task Demonstrations}
\label{sec:appendix_200}

\begin{figure*}[ht]
    \centering
    \includegraphics[width=0.325\linewidth]{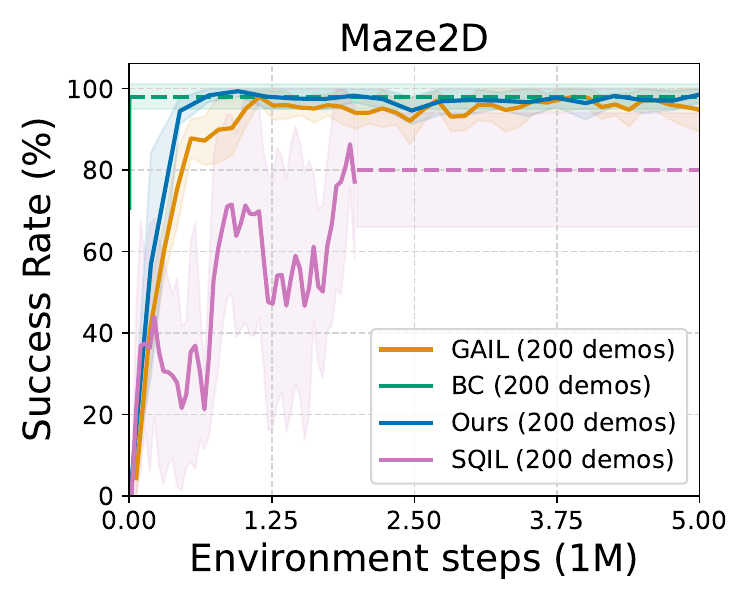}
    \includegraphics[width=0.325\linewidth]{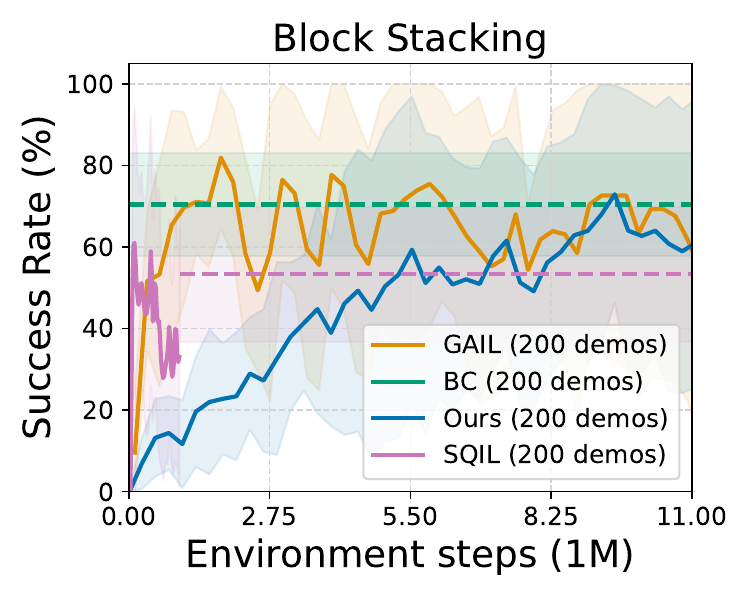}
    \includegraphics[width=0.325\linewidth]{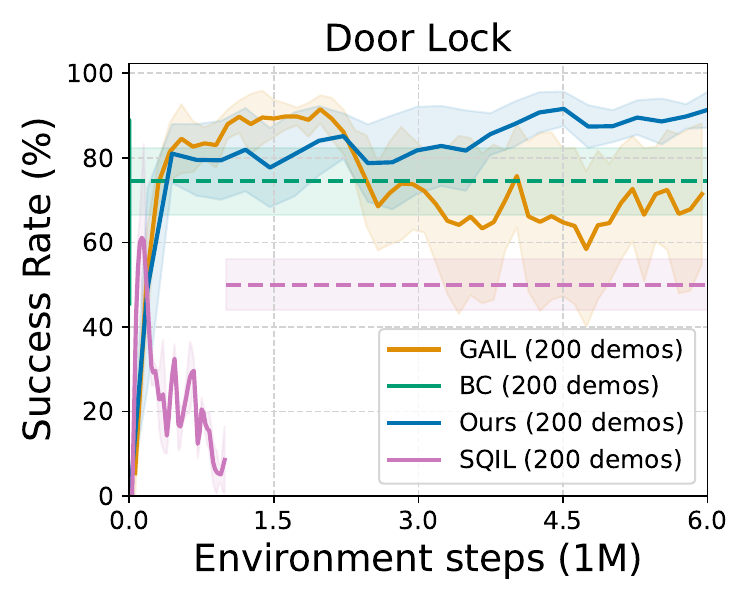}
    \includegraphics[width=0.325\linewidth]{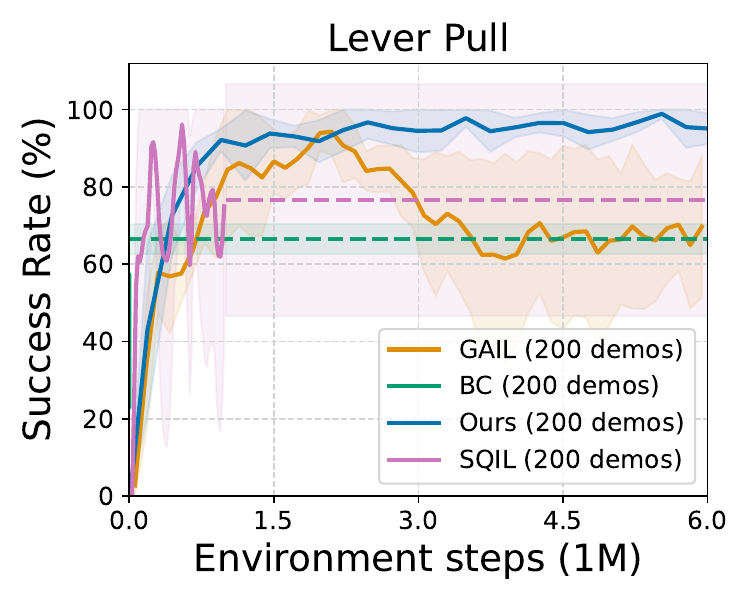}
    \includegraphics[width=0.325\linewidth]{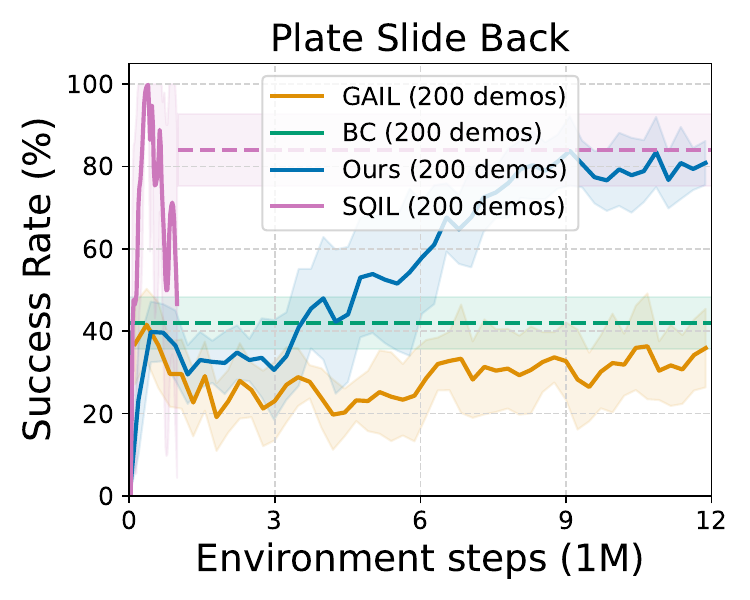}   
    \caption[]{Comparison of \M~with traditional IL and IRL approaches in a standard IL setting with ample demonstrations.
    }
    \label{fig:200_demos}
    \vspace{-10pt}
\end{figure*}

In Figure~\ref{fig:200_demos}, \M~performs competitively with traditional IL and IRL methods, confirming that \M's reward function remains sound when sufficient task-specific data is available.
As expected, however, \M~no longer exhibits a clear advantage, since dense target-task demonstrations sufficiently cover the expert trajectory distribution.
For SQIL, we train under a simplified intra-task variation setting where only the object initial position changes, as SQIL performs poorly under larger intra-task variations even with abundant demonstrations.
These results further demonstrate that GAIL and SQIL are strong IRL baselines, and  their degradation  in our  main experiments arises from limited target demonstrations and substantial intra-task variations rather than inherent methodological weakness. 

\begin{figure}[]
    \centering

    \includegraphics[width=0.32\linewidth]{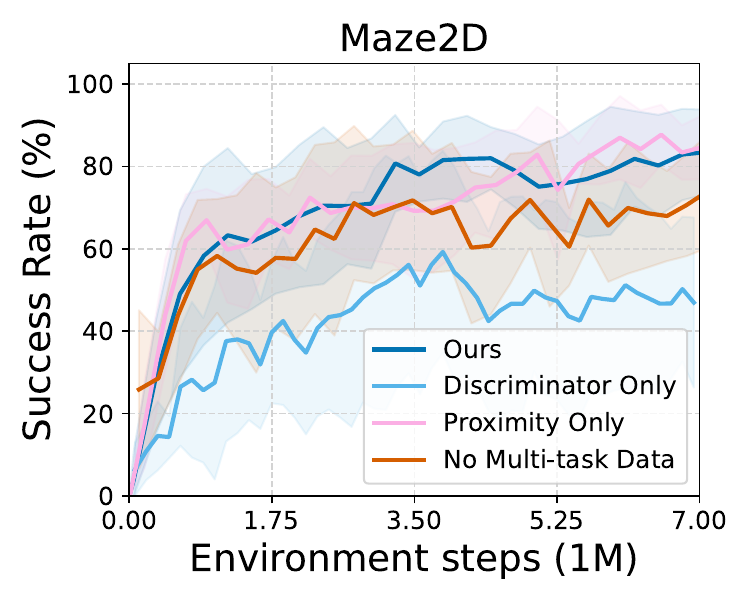}
    \includegraphics[width=0.32\linewidth]{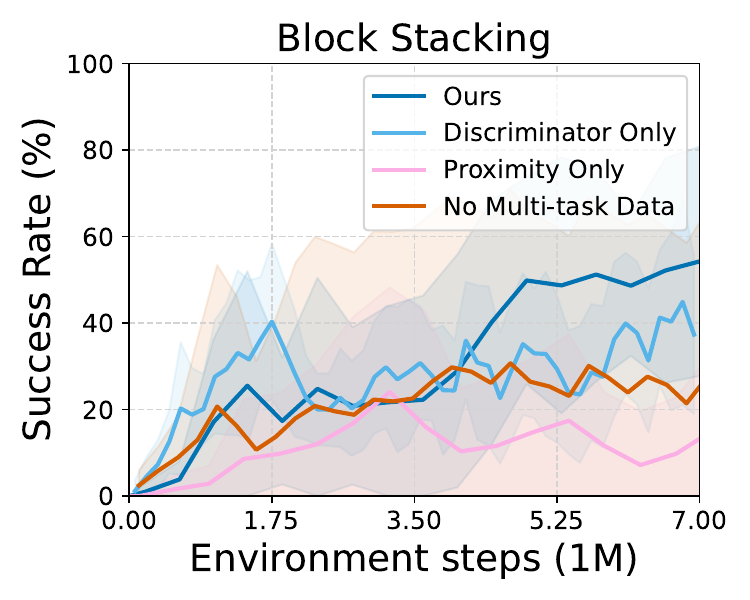} 
    \includegraphics[width=0.32\linewidth]{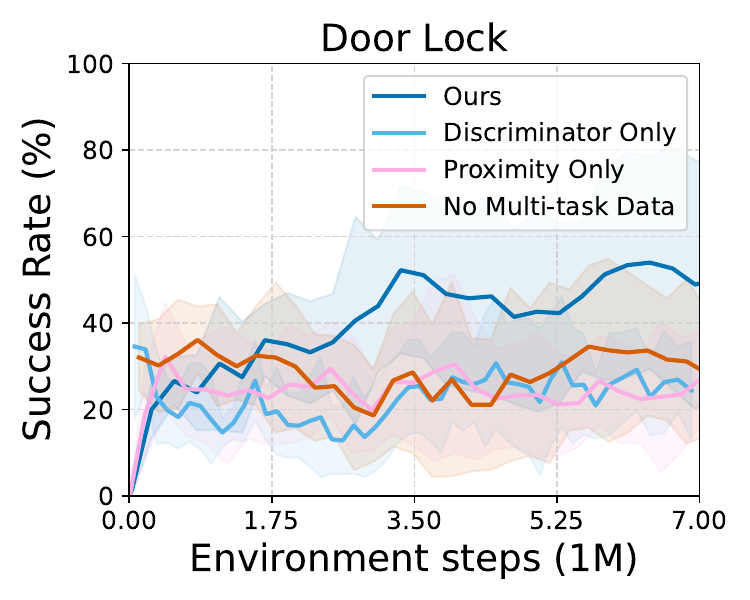}
    \includegraphics[width=0.32\linewidth]{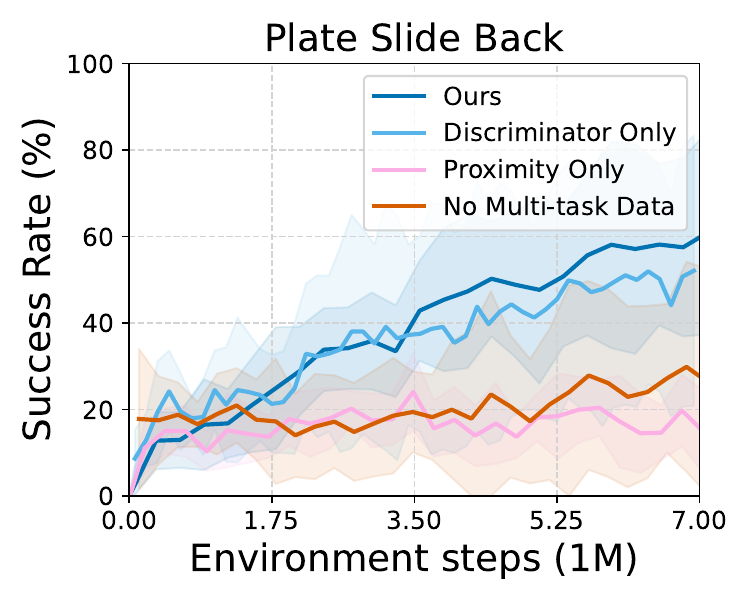}
    \includegraphics[width=0.32\linewidth]{figures/results/ablations/lever-pull-v2_ablation_success_legend.pdf}
    \caption[]{ Ablations over more tasks in supplement to Figure ~\ref{fig:ablation}
    }
    \label{fig:appendix_ablations}
\end{figure}

\begin{figure*}[t]
    \centering
    \includegraphics[width=0.325\linewidth]{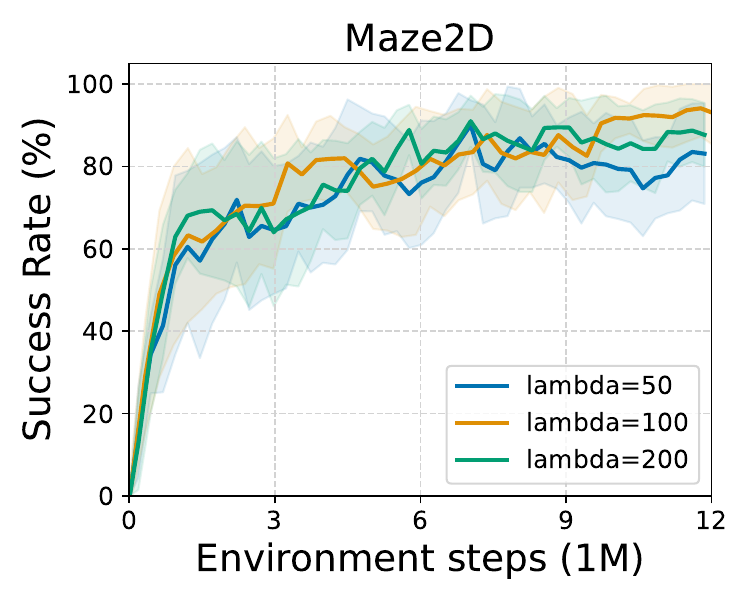}
    \includegraphics[width=0.325\linewidth]{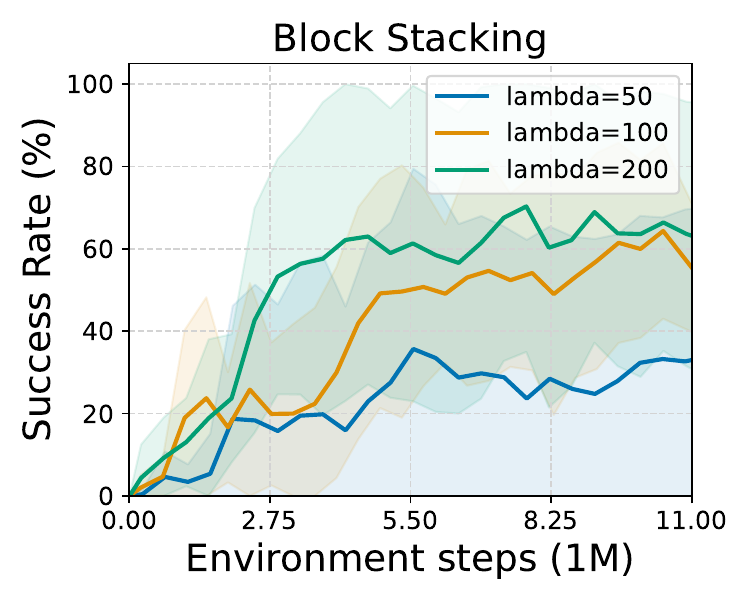}
    \includegraphics[width=0.325\linewidth]{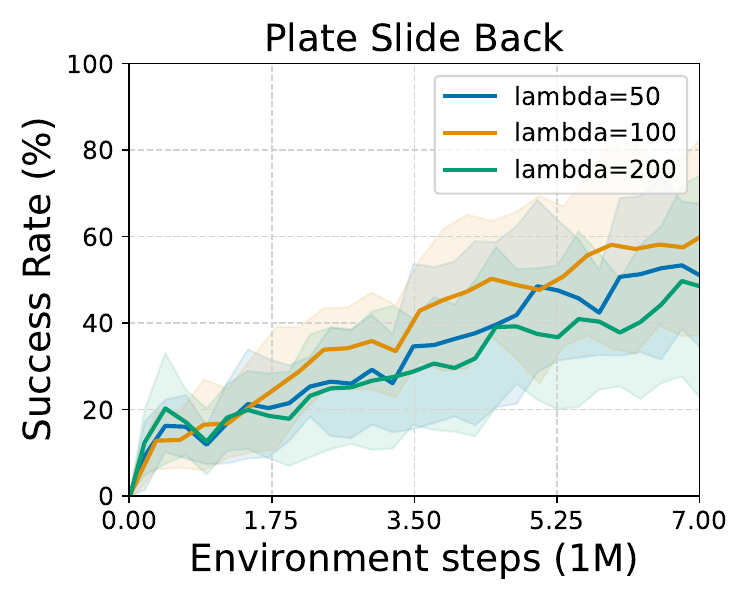}
    \includegraphics[width=0.325\linewidth]{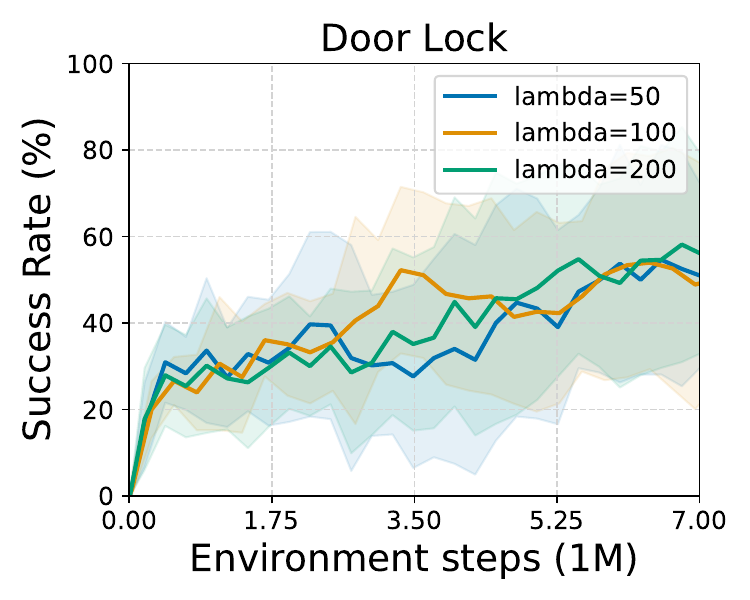}
    \includegraphics[width=0.325\linewidth]{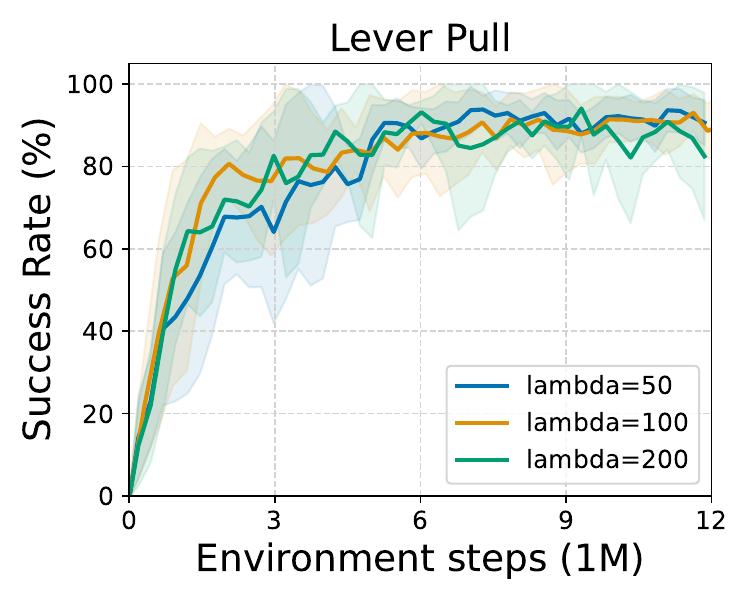}
    
    \caption[]{Analysis on lambda $\lambda$, the coefficient of the proximity reward, in more tasks.
    }
    \label{fig:appendix_lambda}
    \vspace{-10pt}
\end{figure*}

\begin{figure*}[t]
    \centering
    \includegraphics[width=0.325\linewidth]{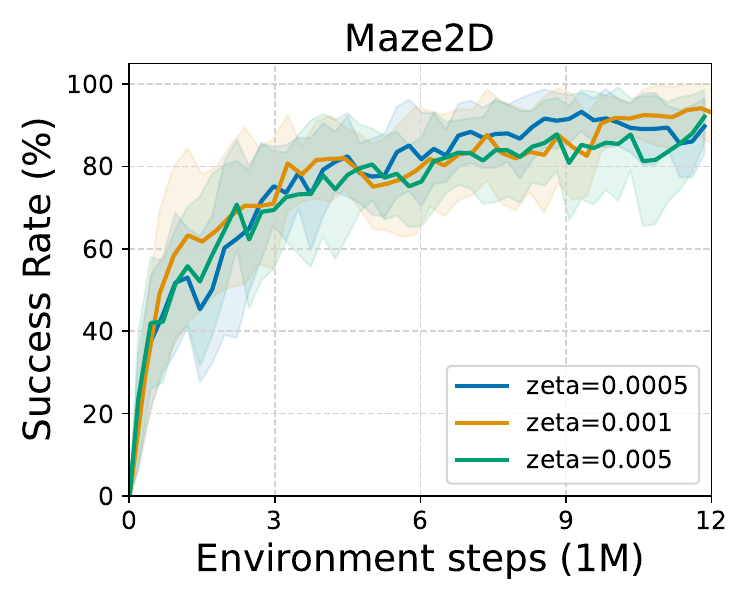}
    \includegraphics[width=0.325\linewidth]{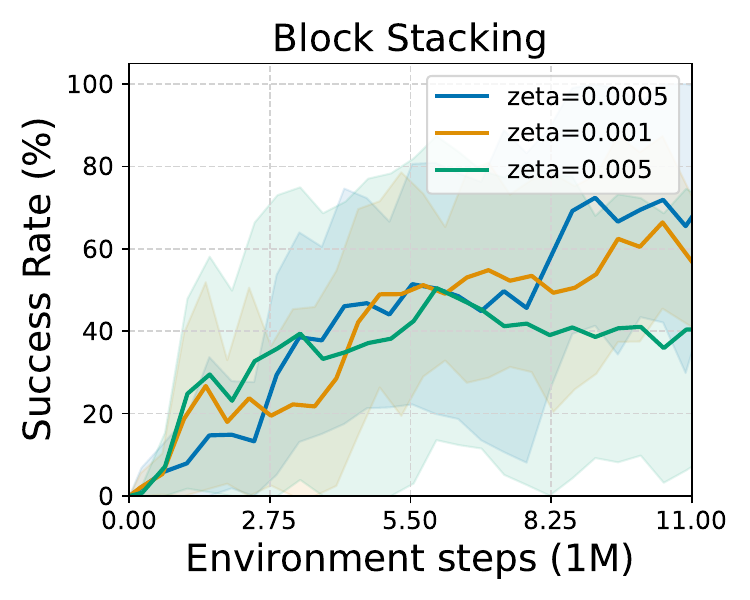}
    \includegraphics[width=0.325\linewidth]{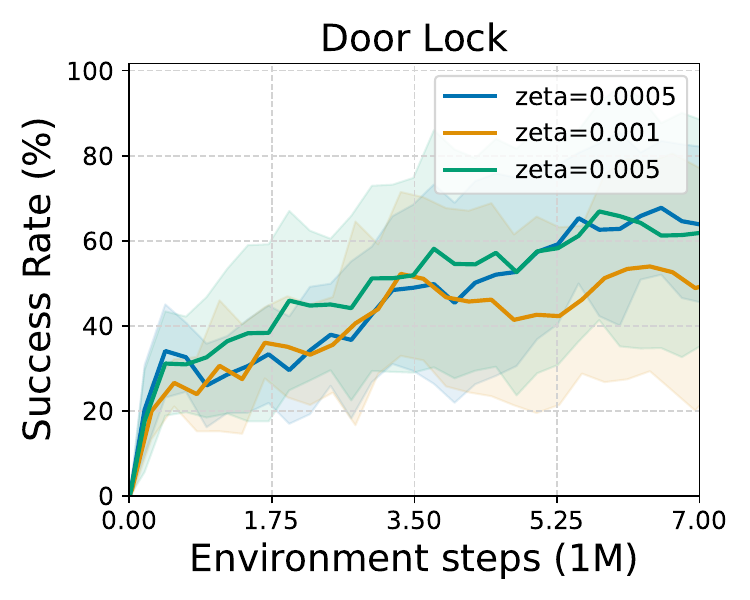}
    \includegraphics[width=0.325\linewidth]{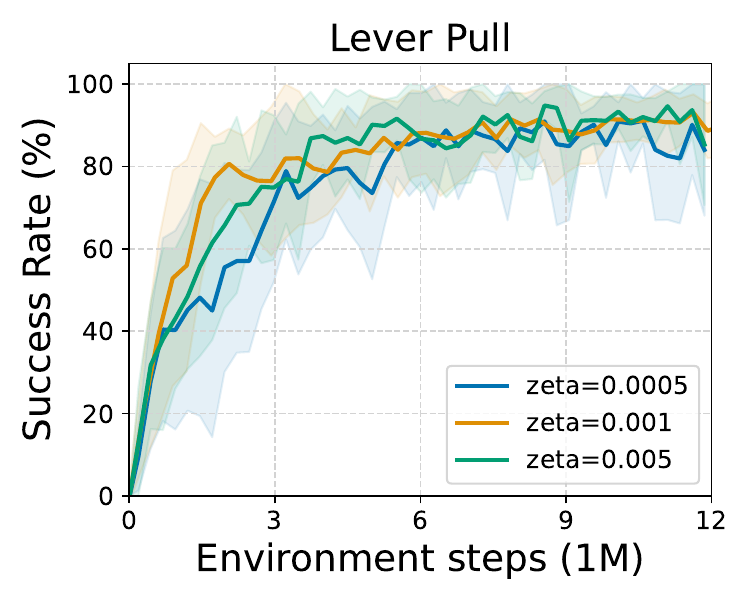}
    \includegraphics[width=0.325\linewidth]{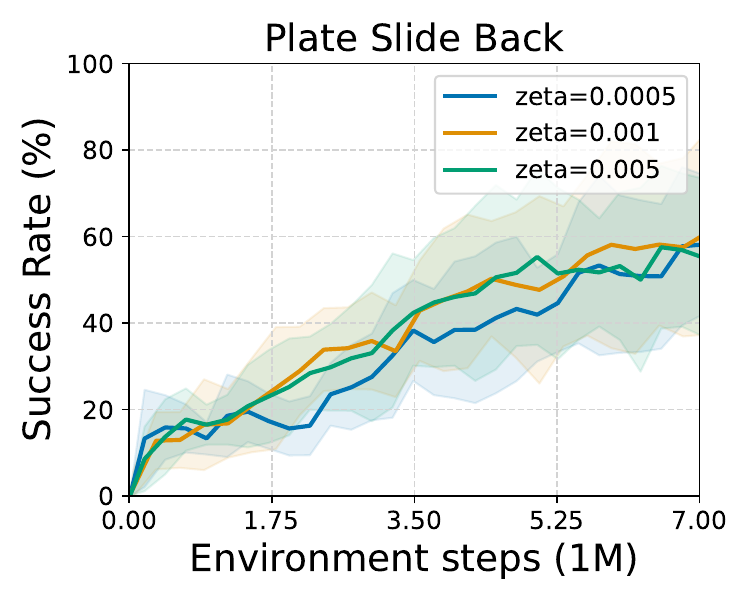}
    
    \caption[]{Analysis on zeta $\zeta$, the proximity timestep factor, in more tasks.
    }
    \label{fig:appendix_gamma}
    \vspace{-10pt}
\end{figure*}

\subsection{Ablation Results on More Tasks}
\label{sec:appendix_ablation}
\textbf{Reward Components.} Figure~\ref{fig:appendix_ablations} contains results ablating the two components of \M's reward function: the multi-task discriminator and proximity function. Across additional tasks,  we see the same trend that combining both components consistently outperforms either component alone. 

\textbf{Multi-Task Dataset.} We also evaluate our method with and without the multi-task demonstrations(Figure~\ref{fig:appendix_ablations}). 
To implement, we train the multi-task discriminator without the multi-task dataset, thus the positive example is only the pairs from the target task.  
We see that the ``No Multi-task Data'' yields performance comparable to the ``Proximity Only'' ablation in most environments, indicating that the discriminator’s primary benefit stems from leveraging multi-task demonstrations.

\textbf{Hyperparameters.}  We further ablate the proximity reward weight $\lambda$ ( Figure~\ref{fig:appendix_lambda}) and timestep factor $\zeta$ (Figure~\ref{fig:appendix_gamma}). Performance remains stable across a broad range of values for both hyperparameters, suggesting that \M~does not require careful fine-tuning. In practice, robust performance is achieved when $\lambda$ is on the same order as the discriminator reward and $\zeta$ is scaled relative to the inverse of the maximum episode length.  
Figure~\ref{fig:appendix_lambda} contains results ablating the proximity reward coefficient $\lambda$ for additional tasks.  We see the same trend here that performance is relatively robust to a range of $\lambda$'s within one order of magnitude, with the exception being Block Stacking at $\lambda=50$. 


Figure~\ref{fig:appendix_gamma} demonstrates similar robustness to $\zeta$ across tasks. 
These findings further support the hyperparameter stability discussed in Section~\ref{sec:mpirl_hyperparameters}.

\subsection{Effect of the Proximity Reward}
\label{sec:appendix_prox}
We analyze the contribution of the proximity component in \M’s reward and its interaction with the multi-task discriminator.  The proximity function estimates the number of steps away a state is from an expert state, providing informative feedback in non-expert regions.  Theoretically, this same reward can be combined with traditional IRL methods that do not use multi-task data.  

\begin{figure*}[ht]
    \centering
    \includegraphics[width=0.325\linewidth]{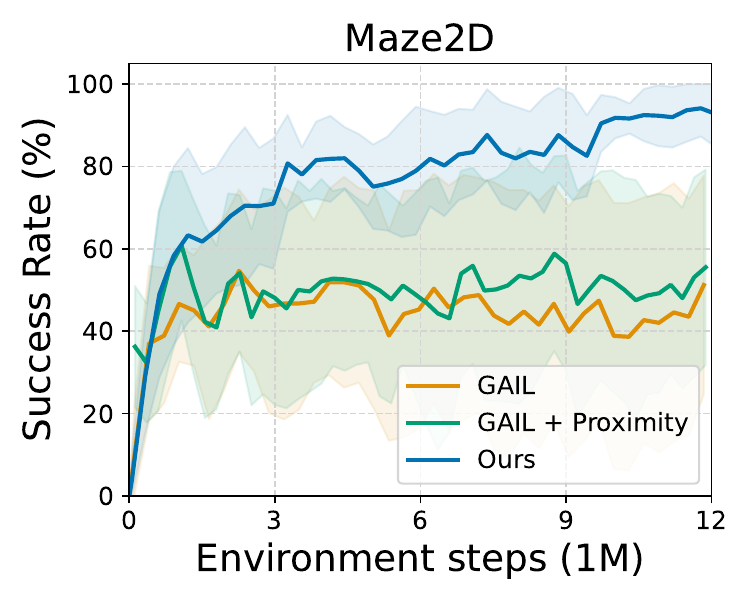}
    \includegraphics[width=0.325\linewidth]{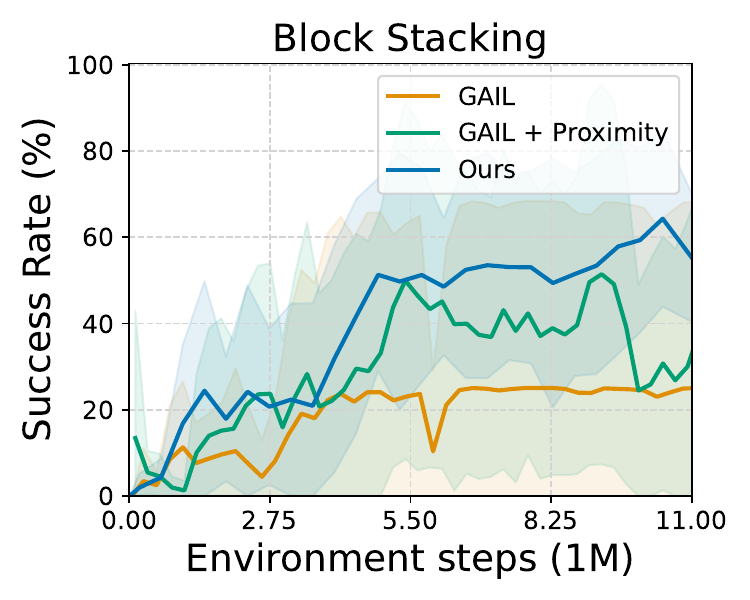}
    \includegraphics[width=0.325\linewidth]{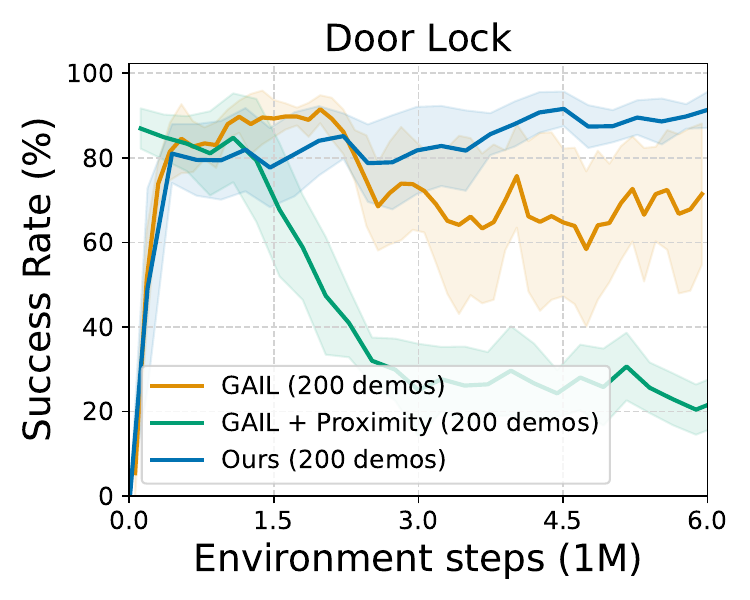}
    \includegraphics[width=0.325\linewidth]{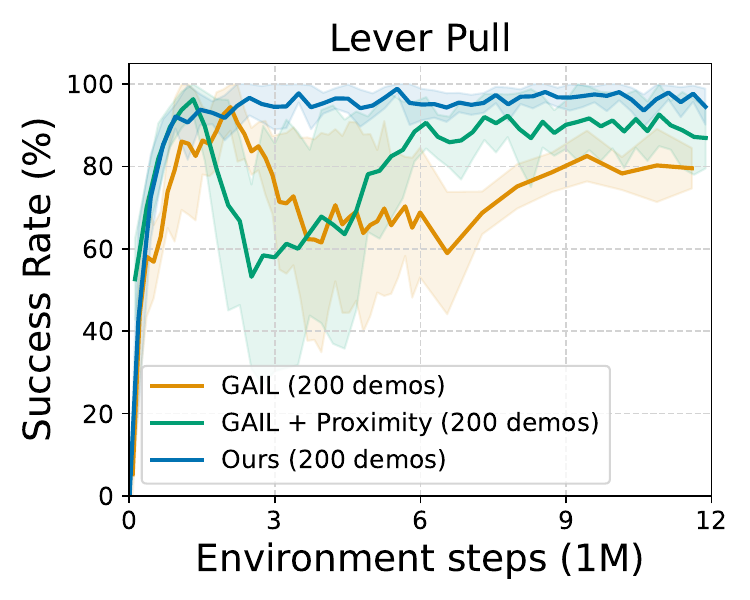}
    \includegraphics[width=0.325\linewidth]{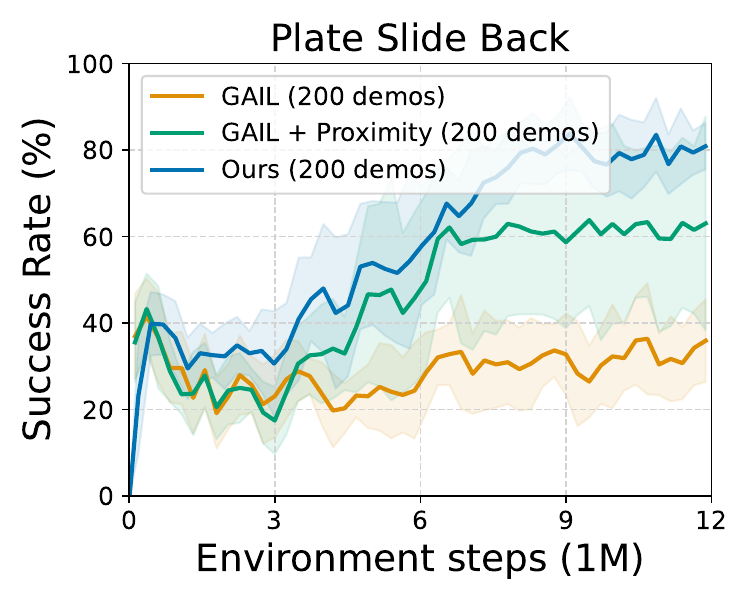}   
    \caption[]{Analysis of our proximity reward by combining it with GAIL without multi-task data (GAIL + Proximity). 
    }
    \label{fig:prox}
    \vspace{-10pt}
\end{figure*}

To evaluate its standalone benefit, we augment GAIL with the proximity reward (Figure~\ref{fig:prox}).  In 3 out of 5 tasks, this improves GAIL's performance and stability, with gains of up to 20\% in Plate Slide Back, supporting the usefulness and soundness of the proximity reward on its own. 

For the FactorWorld tasks, 200 target demonstrations were required for ``GAIL + Proximity'' to yield consistent improvements.   
For Lever Pull and Door Lock, we additionally tuned the re-labeling hyperparameter $c_{thresh}$ ($0.4$ and $0.6$, respectively). These changes improve discriminator generalizable and increase re-labeled expert states.  
Across all tasks, \M~outperforms ``GAIL + Proximity'' , validating the benefit of the multi-task discriminator, even in data-rich settings.



\section{Environment Details}
\label{sec:appendix_environment}

\subsection{Maze2D}
    We base our implementation on the Maze environment from the D4RL benchmark \citet{fu2020d4rl}. As show in Figure \ref{fig:env_maze}, there are four balls placed in fixed locations, resulting in four tasks. The starting positions of the agent are randomly sampled. The state space is the agent's position, velocity, and positions of four balls, and then outputs an x- and y-velocity to navigate in the maze. Episodes have a horizon of 1500 timesteps. For the target task we use two demonstrations, and for the multi-task dataset we use 200 demonstrations for each of the remaining three tasks, all gathered by a planner-based policy provided in \citet{pertsch2021accelerating}.

\subsection{Block Stacking}

    We use the implementation from \citet{pertsch2021accelerating}, there are five blocks on the ground with five different colors. The five block starting positions are randomly generated. In each task, the agent aims to pick up a block with color X and place it on a block with color Y (X and Y are two different colors selected from five colors). Different tasks have different pick-place colors. The state space contains the gripper's position, opening angle, velocity, and the position of the gripper fingers. It also includes the position and orientation of the block in quaternions. The action space consists of an (x, z)-displacement and a continuous action representing the degree of the robot gripper's opening. We collect 200 demonstrations for each task using a planner from \citet{pertsch2021accelerating} and use 25 demonstrations for the target task. The target task is to stack the purple block on top of the blue block. 
    The three tasks in the multi-task demonstration dataset are: purple on top of green, black on top of blue, and green on top of white. Episodes have a horizon of 500 timesteps.

    This task is very unforgiving: dropping a block prematurely drives the policy out of distribution, from which recovery is difficult. We hypothesize that our proximity reward mitigates such errors by penalizing those transitions more than other less harmful non-expert behaviors.
    
\subsection{FactorWorld}
    We utilize the implementation provided by \citet{xie2024decomposing}, which extends the Meta-World benchmark \citep{yu2020meta} by introducing various factors of variations. In our experiments, we incorporate variations in object position, table position, and arm position, and include distractor objects with diverse initial positions and shapes. The agent observes in state space, the 3D position of its end effector, how open its gripper is, the 3D positions of the one or two objects on the tabletop, table position, the goal position, and its previous state.  The action space is the end effector position delta along with the normalized torque input to the gripper.  We evaluate performance on seven tasks from the benchmark, using between 2 and 25 demonstrations for each task (Table~\ref{tab:factorworld_demos}).  Since these tasks vary by difficulty, what is considered too few demonstrations varies.  Additionally, we leverage an offline dataset consisting of 10 tasks randomly selected from the following set of 18 tasks, none of which are target tasks: reach, push, pick-place, dial-turn, drawer-close, button-press, peg-insert-side, window-open, sweep-into, basketball, door-close, faucet-open, hammer, handle-press-side, pick-out-of-hole, plate-slide, plate-slide-side, handle-pull. Each of these tasks has 200 demonstrations, collected by Meta-World's open-source hard-coded policies. The maximum number of timesteps per episode is capped at 500.

\begin{table*}[ht]
    \caption[FactorWorld Number of Target  Demos]{FactorWorld Number of Target Demos}
    \centering
    \begin{tabular}{l c c c c c c c}
        \toprule
        {Task} & 
            Drawer & Door & Door & Plate & Door & Lever & Button \\
          & Open & Lock & Unlock & Slide Back & Open & Pull & Press Wall \\
        \midrule
        \# Demos & 5 & 10 & 5 & 5 & 2 & 25 & 10 \\

        \bottomrule
    \end{tabular}
    \label{tab:factorworld_demos}
\end{table*}

\subsection{Minigrid}
\label{env:minigrid}
There are four tasks, each requiring the agent to reach a different corner of the room, with the target task being to reach the bottom-left corner. The agent's start position is randomly initialized.
The environment’s discrete action space includes movement directions: up, down, left, and right. The state space includes the maze layout, agent position, and direction. A single expert demonstration is provided for the target task.  We provide 200 demonstrations for each of the remaining three tasks
For interpretability, we consider a state-only discriminator and define the proximity function such that all expert states have proximity value $1$, yielding high proximity within the expert distribution and progressively lower values as the agent deviates. 


    
    

\section{Implementation Details}
\label{sec:appendix_implementation_details}
We use the robot learning code base from \url{https://github.com/youngwoon/robot-learning} for basic RL and imitation learning baselines and use default hyperparameters unless otherwise specified. All online methods use PPO~\citep{schulman2017proximal} as the RL algorithm except SQIL which uses the off-policy algorithm SAC~\citep{haarnoja2018sac}.  For all methods, we initialize the policy with a BC trained policy and add an auxiliary BC loss to the policy loss function using Equation~\ref{eqn:bc_loss}.  We do not use BC to initialize PEMIRL since it infers and conditions on a context variable that cannot be pretrained with BC.  We detail our own implementations of each method below.

\begin{equation}
    \label{eqn:bc_loss}
    L_{MSE} = \mathbb{E}_{(s, a) \sim \mathcal{D}_{target}}  \|a - \pi(s)\|^2 
\end{equation}

\subsection{SQIL}
\label{sec:appendix_sqil}
We implement SQIL using the resources from~\citet{reddy2020sqil} and use SAC as the off-policy RL algorithm. It gives sparse rewards (i.e., $+1$ only to transitions inside expert demonstrations and $0$ elsewhere). To incorporate the other task data, we add it to the training data with labeled rewards of $0$.  For each batch of training data, we sample 50\% from target task demonstrations, 40\% from the policy replay buffer, and 10\% from the multi-task demonstrations.  This addition can provide better coverage of the environmnet especially early on in training.

We run SQIL until convergence, which often happened more quickly than the other methods because SAC tends to be more sample efficient than PPO. SQIL requires an off-policy RL algorithm.  While our method could also use SAC, in practice, we found the generative adversarial training for the multi-task discriminator to be more stable with PPO.

\subsection{GAIL}
We train GAIL by treating the multi-task demonstrations as additional negative examples for the discriminator, in addition to the standard online policy samples. 

\subsection{MT-AIRL}
We implement MT-AIRL by training a multi-task demonstration-conditioned discriminator using the same network architecture and training procedure as our multi-task discriminator.  We train MT-AIRL on $\mathcal{D}_{target} \cup \mathcal{D}_{multi}$ using demonstration trajectories and expert state-action tuples from the same task as positive classification examples and state-action tuples from different tasks as negative examples.

\subsection{DVD}
We implement DVD and adapt the video-discriminator from a non-Markovian reward function to a state-action based reward function.  Specifically, we input a demonstration trajectory including actions, and  state-action tuple, and predict whether or not that state-action tuple exhibits expert behavior for the demonstrated task.  Similar to our multi-task discriminator, we train DVD on $\mathcal{D}_{target} \cup \mathcal{D}_{multi}$ using trajectory and state-action tuples from the same task as positive samples and trajectory and state-action tuples from different tasks as negative examples.  We train DVD for 200 gradient steps using batch size of 128 and learning rate of 1e-3 then use it as a reward function to train a policy with online RL.

\subsection{PEMIRL}
We use the implementation from \citet{MHAIRL}, and made the following changes to accommodate our problem setting.  We ensure that sampling between demonstrations of different tasks are balanced so the target task gets sufficient training. Since PEMIRL learns a single policy over all tasks in the meta-training set, we use the same network architectures as the other methods but with double the width (512 hidden dimensions), in order to accommodate the higher capacity.  In the Factorworld tasks, we removed pick-out-of-hole from the list of tasks that can be sampled for the multi-task demonstration set because training in the pick-out-of-hole environment frequently caused unstable simulation and training.  To clarify, agents still received the same number of tasks and demonstrations, but the multi-task demonstrations were sampled out of 17 instead of 18 like the other comparison methods.  

\subsection{GoalPro}
We use the implementation from \citet{lee2021generalizable}, with PPO as the RL algorithm.  
The limited target demonstrations are used by GoalPro to learn its reward function. 
We do not provide the additional multi-task demonstrations, as there is no straightforward way to assign goal-proximity labels for the target task.

\subsection{\M}
\label{sec:mpirl_implementation}
Before starting online training.  
We pre-train $p_{\theta}$ only on the demonstration data by treating $\mathcal{D}_{target}$ as expert states and $\mathcal{D}_{multi}$ as non-expert states and optimizing the same objective Eq.~\ref{eqn:prox}.
 
During online training, we alternative between updating the policy $\pi$, multi-task discriminator $d_{\phi}$, and proximity function $p_{\theta}$, training the policy and multi-task discriminator adversarially while updating the proximity function with current policy samples. Initially, we collect 2000 steps of policy data from the environment, storing it in two separate buffers: the policy replay buffer $\mathcal{D}_\pi$ (for policy data with predicted rewards $\tilde{R}$) and the proximity dataset $\mathcal{D}_{prox}$.  In our implementation, to facilitate balanced sampling of expert states, we maintain two separate buffers for expert and non-expert states that together make up $\mathcal{D}_{prox}$.  Policy samples re-labeled as expert by the multi-task discriminator ($d(s,a) > c_{thresh}$) are added to the expert buffer, along with the target task demonstrations from $\mathcal{D}_{target}$.  When training $p(s)$, we sample two minibatches of non-expert states (one for the maximization objective and one for the triangle inequality constraint) and one minibatch of expert states (for the expert proximity anchoring).  We found that this improves sample efficiency.  We use a sigmoid output activation to cap our proximity values between 0 and 1, which 0 being closest to expert states and 1 being the furthest. 
The policy can be trained with any RL algorithm to optimize this combined reward. 

\subsection{General Hyperparameters}
\label{sec:hyperparameters}

For all environments, we use a learning rate of 3e-4 for policy and 1e-3 for the reward function.  We use PPO with a clip ratio of $0.2$ and a batch size of $128$.  The proximity function is a feedforward network with 2 hidden layers of dimension 256 and tanh activation.  The multi-task discriminator has the same architecture with an added lstm (2 layers, hidden dimension 128) to encode the demonstration trajectory, which is concatenated with the state-action tuple.  The RL policy and critic are feedforward networks with 2 hidden layers of dimension 256 and relu activation.


\subsection{\M~Hyperparameters}
\label{sec:mpirl_hyperparameters}
\M~has two main hyperparameters: the proximity reward weight $\lambda$, the cost of each timestep $\zeta$. While both can be tuned for optimal performance, we provide the best practices for selecting good values based on the environment, without the need for extensive tuning.  The hyperparameters used in our experiments are summarized in Table~\ref{tab:hyperparams}.  


\textbf{Selecting $\lambda$.}  We selected $\lambda = 100$ for all our experiments to match the proximity reward, $\lambda[p(s_t) - p(s_{t+1})]$ to the observed magnitude of the discriminator rewards. As we see in Figure~\ref{fig:appendix_lambda}, \M~is robust to a range of $\lambda$'s within the same order of magnitude, so this hyperparameter is not very sensitive and can be selected without tuning.


\textbf{Selecting $\zeta$.}
$\zeta$ affects how quickly the proximity reward decreases as states get further away from the expert state distribution.  A general rule of thumb is to choose $\zeta$ on the order of 1/Maximum episode length, and we found that 
the results with $\zeta = 0.001$ works well for all of our tasks with an episode length between 500 and 1500. This provides informative differences between states while allowing the model to cover a broad range of proximities. Ablations for $\zeta$ are shown in Figure \ref{fig:appendix_gamma}. We see a performance drop-off in some tasks as $\zeta$ becomes too small, but performance is generally stable between $\zeta=0.0005$ and $\zeta=0.005$.  When $\zeta$ is too small, there is very little change in $p_{\theta}(s_t)$, which can lead to higher variance rewards due to noise or errors in $p_{\theta}$.

\textbf{Tuning $c_{thresh}$.}  In practice, a fixed value of  $c_{thresh}$ in the range of $0.8$-$0.9$ is generally sufficient.

\begin{table*}[ht]
    \caption[\M~ Hyperparameters]{\M~hyperparameters.}
    \centering
    \begin{tabular}{l c c c c}
        \toprule
        Hyperparameter & Maze2D & Block Stacking & FactorWorld  \\
        \midrule
        Proximity Reward Weight $\lambda$ & 100 & 100 & 100 \\
        Proximity Timestep Factor $\zeta$  & 0.001 & 0.001 & 0.001 \\
        $d(s, a)$ threshold $c_{thresh}$ & 0.9 & 0.9 & 0.8 \\
        Number of pretraining epochs & 5 & 100 & 5 \\
        Maximum Episode Length & 1500 & 500 & 500  \\
        \bottomrule
    \end{tabular}
    \label{tab:hyperparams}
\end{table*}

\section{Limitations \& Future Work}
\label{sec:appendix_limitation}
\M~requires a structured multi-task demonstration dataset from the same domain and agent. We hope that incorporating large pre-trained language and vision models can ease these assumptions in the future.  In  addition, while \M~is efficient in the number of demonstrations required, it is still fairly sample hungry when it comes to online training.  Combining our reward function with more sophisticated pre-trained behavior models is a promising direction.  Finally, since our reward function is trained online with the policy, it cannot be reused to train a new policy from scratch.

\end{document}